\documentclass{article}

\usepackage[main, final]{neurips_2026}

\usepackage[utf8]{inputenc} 
\usepackage[T1]{fontenc}    
\usepackage{hyperref}       
\usepackage{url}            
\usepackage{booktabs}       
\usepackage{amsfonts}       
\usepackage{nicefrac}       
\usepackage{microtype}      
\usepackage[table]{xcolor}  
\usepackage{multirow}
\usepackage{graphicx}
\usepackage{amsmath}
\usepackage{enumitem}
\usepackage{subcaption}
\usepackage{algorithm}
\usepackage{algorithmic}
\usepackage{float}
\usepackage{wrapfig}
\title{Imitation from Observations with Trajectory-Level Generative Embeddings}

\author{%
Yongtao Qu\\
UNC Chapel Hill\\
Chapel Hill, NC 27599 \\
\texttt{yongtao@unc.edu}\\
\And
Shangzhe Li \\
UNC Chapel Hill \\
Chapel Hill, NC 27599 \\
\texttt{shangzhe@unc.edu}\\
\And
Weitong Zhang \\
UNC Chapel Hill \\
Chapel Hill, NC 27599 \\
\texttt{weitongz@unc.edu}\\
}

\begin{document}

\maketitle

\begin{abstract}
We consider the offline imitation learning from observations (LfO), where expert demonstrations are scarce and contain only state observations, and the suboptimal policy is far from expert behavior. In this regime, many existing imitation learning approaches struggle to extract useful information from imperfect data since they impose strict support constraints and rely on brittle one-step models.
To tackle this challenge, we propose \textbf{T}rajectory-level \textbf{G}enerative \textbf{E}mbedding (TGE) for offline LfO. TGE constructs a dense, smooth surrogate reward by using particle based entropy estimation to maximize the log-likelihood of expert trajectories in the latent space of a temporal diffusion model trained on offline suboptimal data. By leveraging the structured geometry of the learned diffusion embedding, TGE captures long-horizon temporal dynamics and effectively bridges the gap under severe support mismatch, ensuring a robust learning signal even when offline data is distributionally distinct from the expert.
Empirically, the proposed approach consistently matches or outperforms prior offline LfO methods across a range of D4RL locomotion and manipulation benchmarks.
\end{abstract}

\section{Introduction}
Imitation learning (IL) has demonstrated remarkable success in enabling agents to acquire complex behaviors across various domains, from robotic manipulation~\citep{florence2022implicit,chi2023diffusion,brohan2023rt1} to autonomous navigation~\citep{codevilla2018end,bansal2019chauffeurnet}. 
While standard IL methods can effectively recover high-performance policies by supervising explicit action labels, applying these techniques to real-world scenarios poses substantial challenges. 
A primary bottleneck is the absence of explicit action information in many natural data sources, such as video demonstrations or third-person recordings, which precludes direct supervision of state-action mappings~\citep{torabi2018behavioral, stadie2017third, liu2018imitation}. 
This has led to the emergence of Learning from Observations (LfO), where the agent must infer desirable behaviors from state-only expert trajectories. 

To recover the underlying actions, 
many of the existing LfO methods attempt to estimate the inverse dynamics or auxiliary inference mechanisms~\citep{hanna2017grounded,kidambi2020morel}. However, an equally critical challenge lies in the quality of the offline data itself. As large-scale offline datasets inevitably contain suboptimal or inconsistent behaviors arising from human error or task ambiguity, directly applying behavioral cloning to such datasets often leads to poor performance. Effectively leveraging these imperfect demonstrations therefore requires algorithms that can robustly filter noise and distill informative behavioral priors, without being misled by non-expert trajectories.

Existing LfO approaches face two structural limitations: surrogate-reward methods based on adversarial discriminators are unstable over long horizons ~\citep{zolna2020offline,ross2011reduction}  and distribution-matching methods rely on density ratios~\citep{ho2016generative, ma2022versatile} that degenerate when offline data has limited support over the expert distribution (see Sec.~\ref{sec:offline IL works}).
These limitations raise a central question for learning from observations: 
\begin{center}
    \emph{How can we extract a reliable imitation signal from imperfect offline data under support mismatch?}
\end{center}

To address these limitations, we propose \textbf{Trajectory-level Generative Embeddings (TGE)}, a novel framework for offline LfO that bypasses the structural reliance on support coverage inherent in density ratio estimation. Instead of relying on unstable density ratios, TGE leverages the smooth representation learned by a trajectory-based diffusion model such as diffuser~\citep{janner2022planning}. By encoding trajectories into this generative latent space, we construct a dense surrogate reward signal that bridges regions of sparse support. 
Specifically, we employ a logarithmic distance kernel for reward estimation inspired by the particle-based entropy estimation~\citep{liu2021behavior, mutti2020policy, singh2003nearest}, which ensures that maximizing the surrogate reward using standard offline RL algorithms, such as IQL~\citep{kostrikov2021offline} or ReBRAC~\citep{tarasov2023revisiting}, is essentially maximizing the log-likelihood of the expert state distribution.
Notably, this design of the surrogate reward also ensures an informative signal throughout the suboptimal data, preventing the reward collapse issues common in prior distance-based methods.

Our main contributions are summarized as follows:
\begin{itemize}[leftmargin=*, topsep=0pt,itemsep=0pt,partopsep=0pt,parsep=0pt]
\item We introduce a trajectory-level representation learning approach based on a generative planner, demonstrating that the resulting latent embeddings are naturally separable and can distinguish between suboptimal and expert-like data in offline datasets without explicit discriminative training. To our knowledge, this is the first use of representations from generative planner in imitation learning. \looseness=-1
\item We derive a reward formulation defined over the trajectory-level latent embedding, which approximately minimizes the particle-based cross-entropy between the policy and expert occupancies, enabling the use of any offline RL methods for imitation learning with observation-only demonstration.
\item Empirically, we identify and analyze a structural failure mode shared by existing commonly used DICE-based methods under support mismatch, and show empirically that TGE remains robust precisely in the regimes where density-ratio methods collapse, matching or outperforming state-of-the-art offline LfO baselines across diverse benchmarks.
\end{itemize}

\section{Related work}
\subsection{Offline imitation learning}
\label{sec:offline IL works}
Offline imitation learning (IL) aims to recover an expert policy from a fixed dataset of demonstrations without access to online environment interactions. Behavioral Cloning (BC)~\citep{pomerleau1989alvinn}, treats this as a supervised regression problem but suffers from distribution shift and compounding errors on out-of-distribution states~\citep{ross2011reduction}. Distribution-matching approaches mitigate by minimizing the divergence between the learner and expert occupancy measures~\citep{kostrikov2020imitation, garg2021iq}.

Learning from Observations (LfO) arises as a more challenging subset where the expert demonstrations lack action labels. Offline LfO approaches generally fall into two distinct paradigms:

\noindent \textbf{Occupancy Matching.} This class of methods formulates imitation as an optimization problem, solving for the state-occupancy ratio $w(s) \approx d^{\pi_E}(s) / d^{\mathcal{D}_\mu}(s)$ between the expert and offline suboptimal data distributions. Methods like SMODICE~\citep{ma2022versatile} and LobsDICE~\citep{kim2022lobsdice}, which minimize the KL-divergence, are often sensitive to the distribution of the offline buffer. Recent variants like PW-DICE~\citep{yan2024offline} introduce Wasserstein geometry to relax the strict support constraints, while DILO~\citep{sikchi2025dual} utilizes a dual formulation of the $\chi^2$-divergence to improve stability. IOSTOM~\citep{pham2025iostom} reformulates the problem as state-transition occupancy matching, avoiding the discriminator pretraining required by earlier DICE variants. Despite these advances, these methods fundamentally rely on estimating density ratios, which can become ill-conditioned when the offline behavioral data has limited support to the expert policy. 

\noindent \textbf{Surrogate Reward Learning.}
Another line of work is to learn a surrogate reward from data, and then optimizes a policy using offline RL algorithms. Broadly speaking, this paradigm extends beyond imitation learning and is central to unsupervised reinforcement learning~\citep{laskin2021urlb}, where methods typically derive intrinsic pseudo-rewards to guide policy acquisition or environment exploration in the absence of external supervision. In the context of offline LfO, ORIL~\citep{zolna2020offline} apply adversarial learning to train a discriminator-based reward, which is notoriously unstable and often produce signals that are difficult to optimize. A more recent non-adversarial alternative, SEABO~\citep{lyu2024seabo}, constructs rewards from nearest-neighbor distances to expert demonstrations via a KD-tree, sidestepping discriminator instability but relying on a fixed observation-space metric that lacks temporal context. TGE falls into this surrogate reward paradigm but circumvents both limitations by deriving a dense, theoretically grounded reward from the latent space of a trajectory-level generative planner trained on the offline dataset. 

\subsection{Diffusion models in reinforcement learning} 
Diffusion models have been widely adopted in RL as policies, trajectory planners, value regularizers, and world models~\citep{janner2022planning,ajay2022conditional,chi2025diffusion,wangdiffusion,ding2024diffusion,ho2020denoising,alonso2024diffusion}.
In the context of diffusion models in RL, our work focuses on a distinct application in using diffusion models for metric learning. 
While methods like Stable Rep~\citep{tian2023stablerep} or Diffusion Reward~\citep{huang2024diffusion} similarly derive learning signals from generative models, they typically rely on static visual semantics from video or image models. In contrast, TGE exploits the trajectory-level latent embedding of a temporal diffusion encoder to capture system dynamics, augmenting the reward-free offline dataset that enables robust imitation without the computational cost of generative sampling during downstream offline RL training.

\subsection{Representation learning in reinforcement learning}
\label{sec:rep_learning}
Representation learning is central to scaling RL to high-dimensional observation spaces by extracting compact embeddings that filter task-irrelevant noise. 
Early approaches largely relied on reconstruction-based auxiliary tasks with world models or autoencoders to compress sensory inputs into latent states that capture the factors of the environment~\citep{ha2018world}.
Subsequent research adopted contrastive learning and bi-simulation metrics to group states with similar transition dynamics~\citep{laskin2020curl,zhang2021learning} to improve the model robustness to task-irrelevant details. 
However, these approaches typically learn the representation for single-stage observation. To tackle this, we instead leverage the representation learned from a trajectory diffusion model (i.e., Diffuser~\citep{janner2022planning}). 
These trajectory-level representations provide the temporal context necessary to distinguish behaviors that are visually similar but possess divergent underlying dynamics. Recent works have shown that diffusion models can be viewed as denoising autoencoders where intermediate features serve as robust representations for downstream tasks~\mbox{\citep{fuest2024diffusion}}, suggesting that diffusion training induces semantically meaningful features beyond image generation.

\section{Preliminaries}
\noindent \textbf{Markov Decision Process.} We consider the discounted Markov Decision Process (MDP) setting, defined by the tuple 
$\mathcal{M} = (\mathcal{S}, \mathcal{A}, P, r, \gamma)$, 
where $\mathcal{S}$ and $\mathcal{A}$ denote the state and action spaces, respectively. 
The transition kernel $P(s'|s, a)$ specifies the probability of transitioning to next state $s'$ given the current state $s$ and action $a$ with the reward function defined by $r(s, a)$ and 
$\gamma \in [0,1)$ is a discount factor.
A policy $\pi(a|s)$ defines a distribution over actions conditioned on the state, and the objective of reinforcement learning is to identify a policy that maximizes the expected discounted return
$
    J(\pi)
    = \mathbb{E}_{\pi, P}\!\left[\sum_{t=0}^{\infty} \gamma^t r(s_t, a_t)\right].
$


\begin{figure*}[t] 
\centering
\includegraphics[width=\textwidth]{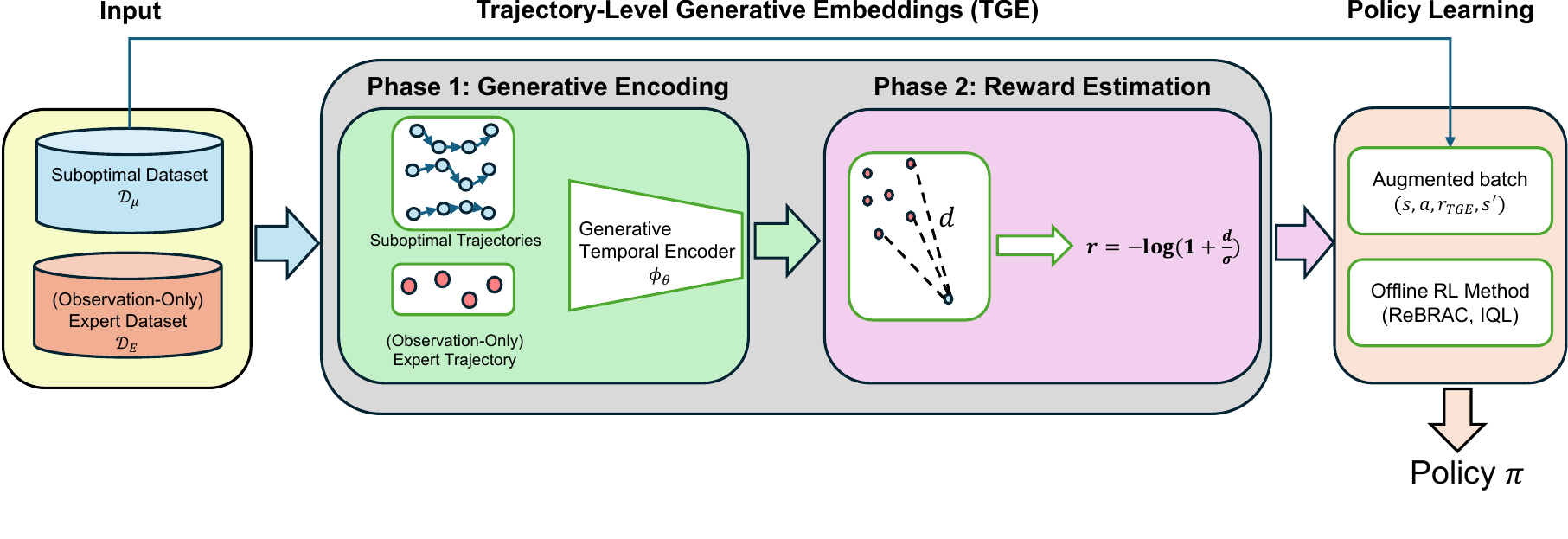}
\vspace{-2em}
\caption{\textbf{Overview of Trajectory-level Generative Embeddings (TGE).} The framework employs a trajectory-level diffusion encoder to map trajectory segments into a latent embedding space. A surrogate reward is then computed using a logarithmic kernel over these embeddings to augment the reward-free suboptimal dataset, enabling offline RL training for policy learning.
}
\vspace{-1em}
\label{fig:pipeline}
\end{figure*}

\noindent \textbf{Offline Imitation Learning from Observations.}
In imitation learning from observation (LfO), the agent is given a small expert dataset containing
only state observations, without any expert actions or reward annotations. In addition, our work
focuses on the offline learning, where the agent cannot directly interact with the environment. Instead, we assume the agent can access an offline dataset composed of suboptimal state--action trajectories, which provides the necessary coverage for learning.

Let the expert dataset containing one complete state-only expert episode be denoted by $\mathcal{D}_E$, and let $\mathcal{D}_\mu$ denote the suboptimal offline dataset consisting of state--action trajectories generated by a mixed behavioral policy $\mu$, where $\mathcal{D}_\mu$ contains a combination of expert and suboptimal trajectories. The agent does not have access to the trajectory labels; that is, it cannot directly distinguish which trajectories are expert and which are suboptimal. This setting is standard practice in offline imitation learning from observations~\citep{sikchi2025dual, yan2024offline}.
 We define the normalized discounted state visitation distribution of a policy $\pi$ as $\rho_\pi(s) = (1-\gamma)\sum_{t=0}^\infty \gamma^t P(s_t=s|\pi)$. Accordingly, let $\rho_E$, $\rho_\pi$, and $\rho_\mu$ represent the distributions induced by the expert policy $\pi_E$, the learned policy $\pi$, and the data-collecting policy $\mu$, respectively. The goal of the learning algorithm is to recover a policy $\pi$ whose behavior closely matches that of the expert policy $\pi_E$.
 
\noindent \textbf{Denoising Diffusion Probabilistic Models.}
Denoising Diffusion Probabilistic Models (DDPMs)~\citep{ho2020denoising} approximate complex data distributions via iterative denoising. The forward process gradually corrupts a clean sample $\mathbf{x}_0 \sim q(\mathbf{x}_0)$ by adding Gaussian noise over $K$ steps according to $q(\mathbf{x}_k \mid \mathbf{x}_{k-1}) = \mathcal{N}\!\left(\sqrt{1 - \beta_k}\, \mathbf{x}_{k-1},\, \beta_k I\right)$, where $\{\beta_k\}_{k=1}^{K}$ is a variance schedule, yielding $q(\mathbf{x}_K) \approx \mathcal{N}(0, I)$. The reverse process $p_\theta(\mathbf{x}_{k-1}|\mathbf{x}_k)$ is parameterized by a network $\boldsymbol{\epsilon}_\theta(\mathbf{x}_k, k)$ that predicts the injected noise.  Training is performed by minimizing a mean-squared error objective:
\begin{align*}
\mathcal{L} = \mathbb{E}_{k,\mathbf{x}_0,\boldsymbol{\epsilon}}
\left[\|\boldsymbol{\epsilon} - \boldsymbol{\epsilon}_\theta(
\sqrt{\bar{\alpha}_k}\mathbf{x}_0 + \sqrt{1 - \bar{\alpha}_k}\, \boldsymbol{\epsilon}, k)\|_2^2
\right],
\end{align*}
where $\bar{\alpha}_k = \prod_{i=1}^k (1 - \beta_i)$. Recent methods extend DDPMs to the trajectory level, treating $\mathbf{x}_0$ as a full state–action sequence to enable long-horizon planning~\citep{janner2022planning,ajay2022conditional}.

\section{Methodology}
We aim to develop a novel offline imitation learning from observations framework that leverages
trajectory-level generative planners as encoders, which enable the algorithm to reason over extended
horizons and incorporate rich temporal structures. To achieve this, we propose \textbf{T}rajectory-level \textbf{G}enerative \textbf{E}mbeddings (TGE), a method that constructs reward signals from the latent embeddings of a generative planner trained to perform density estimation over the suboptimal dataset $\mathcal{D}_\mu$. These trajectory-level latent embeddings provide a discriminative representation capable of distinguishing expert-like and suboptimal trajectories, thereby serving as an informative reward proxy for offline policy optimization on $\mathcal{D}_\mu$. Finally, we train an offline RL algorithm on the entire offline suboptimal dataset $\mathcal{D}_\mu$ using the TGE-derived reward
function. The overall procedure is summarized in Algorithm~\ref{alg:algorithm_pseudocode}, and an overview of the pipeline is illustrated in Figure~\ref{fig:pipeline}.

\subsection{An entropy perspective on distribution matching}
\label{sec:motivation}
The fundamental goal of imitation learning is to recover a policy $\pi$ such that its induced latent state distribution $\rho_\pi$ matches the expert distribution $\rho_E$. We formalize this by seeking to minimize the particle-based cross-entropy $H(\rho_\pi, \rho_E)$. Given $n$ data points $\{z_i\}_{i=1}^n$ sampled from a latent state distribution induced by a policy $\pi$, we leverage non-parametric entropy estimation techniques~\cite{singh2003nearest,liu2021behavior} to express the particle-based cross-entropy approximation with respect to the expert distribution as:
\begin{align*}\textstyle{
H(\rho_\pi, \rho_E) = -\frac{1}{n}\sum_{i=1}^n \log \left( \frac{m}{N_E v_i^m} \right) + b(m) \propto \sum_{i=1}^n \frac{\log v_i^m}{n},
}\end{align*}
where $b(m)$ is a bias correction term depending solely on the nearest neighbor parameter $m$. Here, $v_i^m$ denotes the volume of a hypersphere with radius $\Vert z_i - z_E^m\Vert$, representing the Euclidean distance between a policy datapoint $z_i$ and its $m$-th nearest neighbor in the expert distribution:
\begin{align*}
v_i^m = \frac{\pi^{d_z/2}}{\Gamma(d_z/2+1)} \Vert z_i - z_E^m\Vert^{d_z},
\end{align*}
where $d_z=\vert\mathcal{Z}\vert$ is the dimension of the latent space and $\Gamma$ is the gamma function. To provide a robust learning signal, we further approximate the $m$-th nearest-neighbor log-volume by averaging the per-neighbor log-volumes over the $m$ nearest expert neighbors $\mathcal{N}_m(\mathcal{Z}_E)$, yielding the approximation:
\begin{align}\textstyle{
H(\rho_\pi, \rho_E) \approx \frac{1}{n} \sum_{i=1}^n \frac{1}{m} \sum_{z_E^k \in \mathcal{N}_m(\mathcal{Z}_E)} \log \Vert z_i - z_E^k\Vert^{d_z}. }\label{eq:l1}
\end{align}
This suggests that minimizing the average log-distance between behavioral samples and their expert neighbors can serve as a surrogate for minimizing cross-entropy.

Since $\log \Vert z_i-z_E^k\Vert^{d_z}
= d_z\log\Vert z_i-z_E^k\Vert$, the fixed positive factor
$d_z$ does not affect the minimizer of the unsmoothed
objective and can be omitted.
Following~\cite{liu2021behavior}, we introduce an additive
constant for numerical stability and define the surrogate reward by
\begin{align}\textstyle{
r(z_i) = -\frac{1}{m} \sum_{z_E^k \in \mathcal{N}_m(\mathcal{Z}_E)} \log\left(1 + \Vert z_i - z_E^k \Vert\right), }\label{eq:rwd}
\end{align}

Using this reward, the cross-entropy objective in~\eqref{eq:l1}
can be written as:
\begin{align}
\widetilde{H}(\rho_\pi,\rho_E)
&= \frac{1}{n}\sum_{i=1}^{n}\frac{1}{m}
\sum_{z_E^k\in\mathcal{N}_m(\mathcal{Z}_E)}
\log\left(1+\Vert z_i-z_E^k\Vert\right) \notag \\
&= -\frac{1}{n}\sum_{i=1}^{n}r(z_i) \notag \\
&\approx -\mathbb{E}_{z_i\sim\rho_\pi}[r(z_i)] \notag \\
&= -(1-\gamma)\mathbb{E}_{\pi}
\left[\sum_{h=0}^{\infty}\gamma^h r(z_h)\right].
\label{eq:goal}
\end{align}
The expectation can be approximated by the sample average $-\frac{1}{n}\sum_{i=1}^{n}r(z_i)$.
The equality in the second line follows from~\eqref{eq:rwd}, and the last line uses the definition $\rho_\pi(\cdot)=(1-\gamma)\sum_{h=0}^{\infty}
\gamma^h P_h(\cdot\mid\pi)$.
Since $(1-\gamma)>0$, minimizing the expected counterpart of this smoothed surrogate is equivalent to maximizing the cumulative surrogate reward. Having established this theoretical framework, we describe our method in the following sections.

\subsection{Generative planner learns separable embeddings}
We treat the encoder of the trajectory diffusion planner Diffuser~\cite{janner2022planning} as an unsupervised representation learner. As discussed in Section~\ref{sec:rep_learning}, the denoising objective encourages the model to learn temporally coherent features that are robust to noise, making them ideal for imitation learning. Specifically, as outlined in Line 1 of Algorithm~\ref{alg:algorithm_pseudocode}, we train the diffusion noise model on the suboptimal dataset $\mathcal{D}_\mu$ using the objective
\begin{align}
\mathcal{L}(\theta)
= \mathbb{E}_{\tau_0 \sim \mathcal{D}_\mu,\, k,\, \boldsymbol\epsilon}
\bigl[\lVert \boldsymbol\epsilon - \boldsymbol\epsilon_\theta(\tau_k, k) \rVert_2^2\bigr],
\label{eq:loss}
\end{align}
where $k \sim \mathrm{Unif}[K]$ denotes the diffusion timestep, $\epsilon \sim \mathcal{N}(0, I)$ is the
Gaussian noise target, and $\tau_k$ is the noisy trajectory obtained by adding noise to the clean trajectory $\tau_0$ at diffusion timestep $k$. Let $\tau \in \mathcal{T}$ denote a trajectory segment of length $H$, where the trajectory space for training is defined as $\mathcal{T} = \mathbb{R}^{H \times (|\mathcal{S}| + |\mathcal{A}|)}$. To construct these segments from $\mathcal{D}_\mu$, we extract overlapping sequences of length $H$ using a sliding window. While the diffusion model is trained on these state--action trajectories, the encoder $\phi_\theta$ is designed to produce a state-only latent embedding. Specifically, during the embedding process, we mask the action dimensions of $\tau$, ensuring the representation space $\mathcal{Z}$ captures temporal state dynamics consistent with the observation-only expert data in $\mathcal{D}_E$. We achieve this by conceptually decomposing the U-Net model $\epsilon_\theta$~\citep{ronneberger2015u} into a temporal encoder $\phi: \mathcal{T} \times \mathbb{R} \to \mathcal{Z}$ and a decoder $\psi: \mathcal{Z} \times \mathbb{R} \to \mathcal{T}$, such that
$\epsilon_\theta(\tau_k, k) = (\psi_\theta~\circ~\phi_\theta)(\tau_k, k)$.
Minimizing the denoising score-matching objective in Equation \eqref{eq:loss} encourages the latent representation $\mathcal{Z}$ to preserve information about the underlying clean trajectory distribution. Consequently, $\mathcal{Z}$ captures temporally coherent features that are stable under perturbations, yielding an embedding space that is more semantically discriminative for downstream similarity-based rewards.

\begin{figure}[t]
\centering
\includegraphics[width=0.9\linewidth]{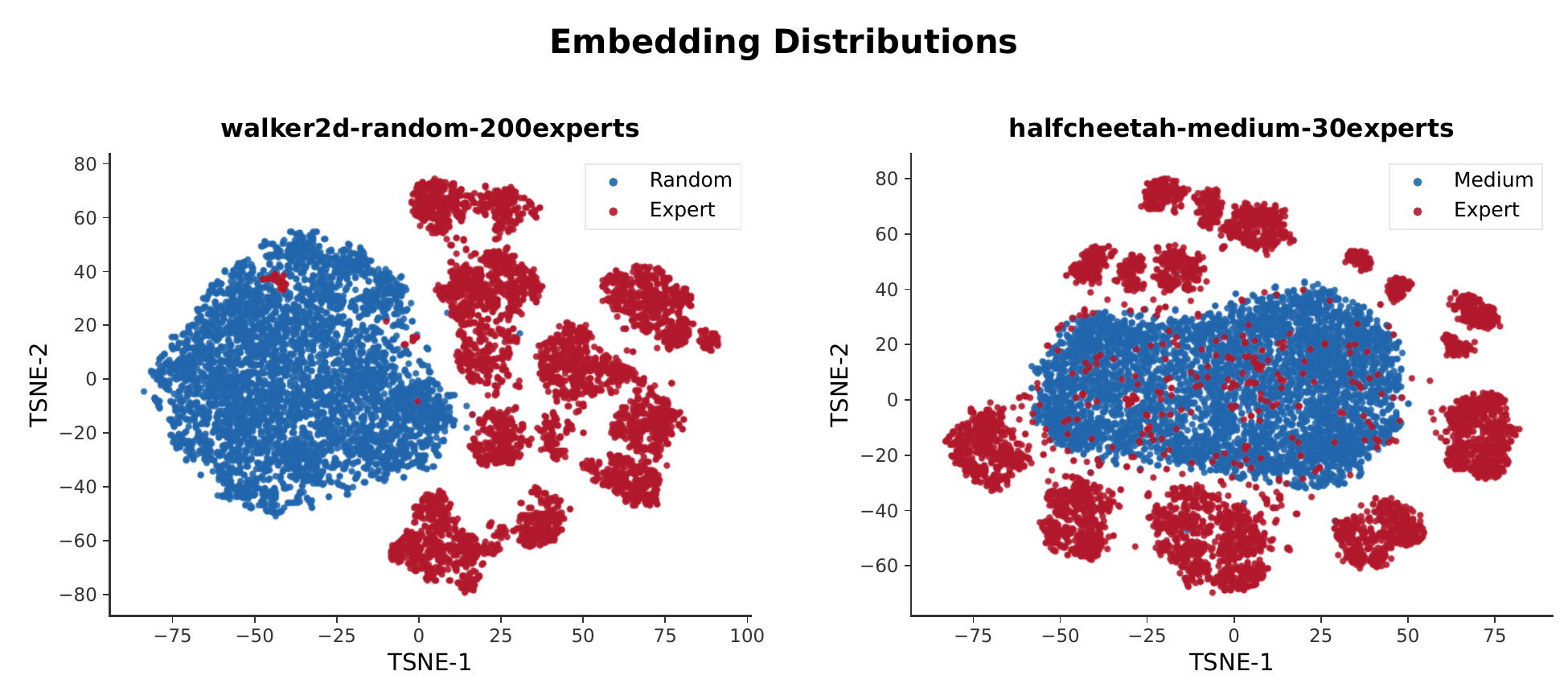}

\caption{\textbf{T-SNE Embedding Visualization.} The latent embeddings produced by the diffuser naturally separate expert transitions from suboptimal ones. Expert ({\color[HTML]{B2182B}red}) and suboptimal ({\color[HTML]{2166AC}blue}) samples form distinct clusters in the embedding space, despite no discriminative supervision.}
\vspace{-1em}
\label{fig:embedding-property}
\end{figure}

Given a clean input trajectory $\tau_0 \sim \mathcal{D}_\mu$, we extract a latent embedding using the encoder. Figure~\ref{fig:embedding-property} illustrates that, even without any explicit discriminative supervision between expert-like and suboptimal trajectories in $\mathcal{D}_\mu$, the encoder-induced embedding $z=\phi_\theta(\tau_0, 0)$ naturally separates expert-like samples from suboptimal ones in the mixed dataset. Given this desirable separation property of the embeddings, it is natural to leverage them for reward
construction by directly measuring an appropriate distance between embeddings as we discussed in Sec.~\ref{sec:motivation}.

\subsection{Reward estimation on trajectory-level embeddings}
\label{sec:reward-estimation}

We now describe the procedure for generating synthetic rewards based on the trajectory-level embeddings induced by the encoder of the generative planner. Given the expert dataset $\mathcal{D}_E$, we extract expert embeddings $\mathcal{Z}_E$ by applying a sliding window with a stride of 1 to the state-only expert episode. Specifically, $\mathcal{Z}_E = \{ \phi_\theta(\tau_{i:i+H}) \mid \tau \in \mathcal{D}_E, 0 \le i \le |\tau|-H \}$, where the action dimensions of the input to $\phi_\theta$ are treated as null for these observation-only segments (Line 2 in Algorithm~\ref{alg:algorithm_pseudocode}). For each state $s_t$ in the suboptimal dataset $\mathcal{D}_\mu$, we consider the trajectory segment starting at time $t$, $\tau_\mu = s_{t:t+H}$, and encode it into a latent vector
$z_t = \phi_\theta(\tau_\mu)$. In the expert embedding space $\mathcal{Z}_E$, we compute the $\ell_2$-norms between $z_t$ and all expert embeddings $z_E \in \mathcal{Z}_E$, and then identify
the top-$m$ nearest neighbors $\mathcal{N}_m(\mathcal{Z}_E)$.

To construct an entropy-based surrogate reward from the suboptimal trajectory segment $\tau_\mu$, we average the kernel-transformed distances between its latent embedding
$z_t$ and its $m$ nearest expert neighbors:
\begin{align}{
r_{\mathrm{TGE}}(s_t)
\!=\! \tfrac{1}{m} \!\!\!\!\sum_{z_E \in \mathcal{N}_m(\mathcal{Z}_E)}\!\!\!\!\!
f\big(\!\tfrac{\| z_t - z_E \|_2}{\sigma}\!\big)\!,
}\notag
\end{align}
where $\sigma$ is a kernel temperature. Note that we normalize all latent embeddings by projecting them onto the unit sphere (dividing each
vector by its $L_2$-norm), ensuring that all embeddings have equal length and allowing for consistent distance computations under the kernel.
 We adopt a logarithmic kernel
$f(d) = -\log(1+d)$. This serves as a proxy for density estimation, consistent with the theoretical derivation linking this metric to particle-based entropy maximization~\cite{liu2021behavior} provided in Section~\ref{sec:motivation}. To justify our theoretical insights, we provide the ablation study in Appendix~\ref{sec:additional-ablation} which shows that our theoretically grounded selection provides the best performance.

\subsection{Downstream policy learning}
Since the original suboptimal dataset $\mathcal{D}_\mu$ consists of state--action transitions without
rewards, the reward estimation procedure described in Section~\ref{sec:reward-estimation}
enables us to annotate each transition with a synthetic reward $r_{\text{TGE}}(s)$, which measures the
similarity of the current state to the expert state distribution. Higher values indicate stronger
expert-likeness, while lower values correspond to more suboptimal behavior.  

With these synthetic rewards, the augmented dataset
$\mathcal{D}'_\mu = \{(s_t, a_t, r_{\text{TGE}}(s_t), s_{t+1})\}$ can be used to train any offline RL method
for policy learning. In our experiments, we adopt two state-of-the-art offline RL algorithms,
IQL~\citep{kostrikov2021offline} and ReBRAC~\citep{tarasov2023revisiting}, and demonstrate that both methods effectively learn high-quality policies when equipped with our synthetic reward signal. Notably, for ReBRAC, we introduce an adaptive behavior cloning mechanism that weights the imitation penalty by the TGE reward to prioritize expert-like transitions during training. Full implementation details are provided in Appendix~\ref{sec:impl}.

\begin{algorithm}[t]
\caption{Trajectory-Level Generative Embeddings}
\label{alg:algorithm_pseudocode}
\begin{algorithmic}[1]
\REQUIRE Offline suboptimal dataset $\mathcal{D}_\mu$, expert dataset $\mathcal{D}_E$, horizon $H$, number of candidates $m$, temperature $\sigma$, and kernel function $f(\cdot)$.
\STATE Train generative planner on $\mathcal{D}_\mu$ to obtain encoder $\phi_\theta$. \label{line:train_encoder}
\STATE Compute embeddings $\mathcal{Z}_{\mu} = \phi_\theta(\mathcal{D}_\mu)$ and $\mathcal{Z}_{E} = \phi_\theta(\mathcal{D}_E)$ on trajectory segments of length $H$. \label{line:compute_embeddings}
\STATE Project all latent vectors to a unit sphere: $z \leftarrow z / \Vert z\Vert_2$. \label{line:normalize}
\FOR{each $z_t \in \mathcal{Z}_{\mu}$}
\STATE Let $d(z_t, z_E) = \|z_t - z_E\|_2$ for all $z_E \in \mathcal{Z}_{E}$. \label{line:calc_dist}
\STATE Identify the top-$m$ nearest neighbors $\mathcal{N}_m(\mathcal{Z}_E)$. \label{line:top_m}
\STATE $r_{\text{TGE}}(s_t) \leftarrow \frac{1}{m} \sum_{z_{nb} \in \mathcal{N}_m(\mathcal{Z}_E)} f\left(d(z_t, z_{nb}) / \sigma\right)$. \label{line:calc_reward}
\ENDFOR
\STATE Run any offline RL algorithm on the suboptimal dataset augmented with synthetic rewards: $\mathcal{D}'_\mu = \{(s_t, a_t, r_{\text{TGE}}(s_t), s_{t+1})\}$. \label{line:offline_rl}
\end{algorithmic}
\end{algorithm}
\subsection{Learning under support mismatch}
\label{sec:analysis}
We highlight a structural difference between TGE and occupancy-matching methods. DICE-based approaches rely on density ratios $w(s) \approx d^{\pi_E}(s) / d^{\mathcal{D}_\mu}(s)$, which typically assume a nontrivial degree of support overlap~\citep{ma2022versatile}. When the offline data poorly covers the expert support, these ratios become ill-conditioned. As $d^{\pi_E}(s) \to 0$, importance weights vanish and supervision collapses. In contrast, the entropy-derived reward $r(s) = -\log(1 + \Vert z_s-z_E \Vert)$ provides a smoothly decaying signal that persists even for states far from the expert support.

The effectiveness of this extrapolation depends on the semantic quality of the underlying metric. Standard distance metrics, such as L2 distance on raw observations, often fail due to state aliasing, where states that are close may be functionally distinct. By leveraging the encoder of a trajectory-level generative planner, the induced latent space is organized according to long-horizon dynamics rather than superficial single-step features. To examine  the contribution of the learned latent geometry, we compare TGE with alternative reward constructions based on raw observations, VAE embeddings, and SEABO~\citep{lyu2024seabo}, using ReBRAC as the shared downstream learner (Section~\ref{app:seabo}). Temporal context also contributes to reward quality, with single-step embeddings substantially degrading
downstream performance (Appendix~\ref{sec:horizon_ablation}).

\section{Experiments}
\subsection{Experimental setup}
\label{sec:exp_setup}

\noindent \textbf{Environments.} We evaluate our method  based on D4RL~\citep{fu2020d4rl} datasets, covering locomotion (MuJoCo: \texttt{hopper}, \texttt{halfcheetah}, \texttt{walker2d}, \texttt{ant}), dexterous manipulation 
(Adroit: \texttt{pen}, \texttt{door}, \texttt{hammer}), and long-horizon multi-task manipulation (Franka Kitchen: \texttt{partial}, \texttt{mixed}).

\noindent \textbf{Expert data.} All methods are additionally provided with an \textbf{observation-only} demonstration $\mathcal{D}_E$ consisting of one single expert episode, which supplies the expert state sequences required.

\noindent \textbf{Suboptimal offline data.} The composition of $\mathcal{D}_\mu$ varies by environment. For Adroit and Franka Kitchen, $\mathcal{D}_\mu$ follows the standard benchmark protocol: \texttt{cloned}/\texttt{human} trajectories for Adroit and \texttt{partial}/\texttt{mixed} datasets for Kitchen, augmented with 30 and 18 unlabeled expert trajectories respectively. For MuJoCo locomotion, we 
consider four mixture settings that vary the level of expert coverage in $\mathcal{D}_\mu$: \emph{expert} (1M suboptimal transitions plus 200 unlabeled expert trajectories), \emph{few-expert} (1M suboptimal 
transitions plus 30 unlabeled expert trajectories), \emph{medium} (1M medium-quality transitions with no expert data), and \emph{medium+random} (1M transitions split equally between medium and random data, no expert). The latter two settings are first introduced in this paper to test performance under severe support mismatch between $\mathcal{D}_\mu$ and the expert. Full mixture details are in Appendix~\ref{sec:mixture-strategy}.

\noindent \textbf{Baselines and Evaluation.} We compare against recent offline learning-from-observations (LfO) methods, including SMODICE~\cite{ma2022versatile}, PW-DICE~\cite{yan2024offline}, DILO~\cite{sikchi2025dual} and IOSTOM~\citep{pham2025iostom}, which have been shown to outperform earlier approaches such as LobsDICE~\cite{kim2022lobsdice} and DemoDICE~\cite{kim2021demodice}. 
We additionally include trajectory-level behavior cloning (BC) as a baseline, trained on the suboptimal data using diffuser~\citep{janner2022planning} without reward guidance. 

We adopt ReBRAC as our default offline RL backbone, paired with adaptive BC weighting (Appendix~\ref{app:bc_weighting}), which yields slightly stronger results than IQL. Training dynamics for both TGE+ReBRAC and TGE+IQL are shown in Appendix~\ref{sec:training-dynamics}, demonstrating that the TGE reward signal transfers across offline RL algorithms. We report the mean and standard deviation of normalized D4RL scores across five seeds. The pseudocode and hyperparameters are in Appendix~\ref{sec:hyperparam}.

\subsection{Experimental results}
\subsubsection{Learning from observations}

\begin{table}[t]
\centering
\caption{\textbf{Main Results.} TGE combined with ReBRAC(highlighted in {\color{cyan}cyan}) is compared against state-of-the-art offline LfO baselines. The results demonstrate that TGE achieves superior performance. The top-performing results and those with overlapping standard deviations are highlighted in \textbf{bold}.}

\renewcommand{\arraystretch}{1.2}
\resizebox{\textwidth}{!}{
\begin{tabular}{|l|l|c|c|c|c|c|>
{\columncolor{cyan!10}}c|c|}
\hline
\textbf{Dataset} & \textbf{Env} & \textbf{Diffuser (BC)} & \textbf{SMODICE} & \textbf{PW-DICE} & \textbf{DILO} & \textbf{IOSTOM} & \textbf{TGE+ReBRAC} & \textbf{Expert} \\ \hline
random+expert & hopper &
1.45$\pm$0.17 & 103.79$\pm$4.76 & \textbf{106.22$\pm$8.21} & \textbf{104.96$\pm$6.52} &
\textbf{109.51$\pm$0.29} & \textbf{109.72$\pm$0.68} & 111.34 \\

& halfcheetah &
0.15$\pm$0.49 & 77.10$\pm$6.72 & 90.63$\pm$1.08 & 89.40$\pm$2.65 &  90.06$\pm$2.12
 & \textbf{93.20$\pm$0.70} & 88.83 \\

& walker2d &
-0.01$\pm$0.04 & \textbf{108.74$\pm$0.72} & \textbf{107.97$\pm$0.94} &
\textbf{108.88$\pm$0.39} & \textbf{107.59$\pm$0.04} &
\textbf{108.68$\pm$0.20} & 106.93 \\

& ant &
9.58$\pm$4.77 & 124.10$\pm$2.19 & 123.15$\pm$8.48 &
122.87$\pm$4.18 & \textbf{128.80$\pm$1.97} &
\textbf{128.26$\pm$2.90} & 130.75 \\ \hline

random+few-expert & hopper &
1.38$\pm$0.06 & 62.04$\pm$17.02 & 89.32$\pm$17.29 &
92.66$\pm$7.42 & \textbf{103.97$\pm$3.48} &
\textbf{106.21$\pm$3.71} & 111.34 \\

& halfcheetah &
0.03$\pm$0.54 & 2.60$\pm$0.64 & 31.16$\pm$32.38 &
48.26$\pm$14.18 & 64.93$\pm$19.04 & 
\textbf{89.70$\pm$1.63} & 88.83 \\

& walker2d &
0.01$\pm$0.03 & 16.97$\pm$35.72 &
97.68$\pm$10.98 & \textbf{107.32$\pm$1.96} & \textbf{107.58$\pm$0.40}
 & \textbf{105.48$\pm$2.52} & 106.93 \\

& ant &
7.25$\pm$0.07 & 32.11$\pm$8.38 &
108.08$\pm$15.99 & 112.50$\pm$7.77 &
\textbf{123.01$\pm$0.82} & 111.46$\pm$4.14 & 130.75 \\ \hline

medium+expert & hopper &
45.52$\pm$4.81 & 57.77$\pm$8.32 & 69.38$\pm$28.15 &
102.52$\pm$5.73 & \textbf{110.54$\pm$0.69} &
\textbf{109.64$\pm$0.25} & 111.34 \\

& halfcheetah &
41.33$\pm$0.49 & 57.27$\pm$2.20 & 60.76$\pm$3.15 &
89.98$\pm$0.61 & \textbf{92.17$\pm$0.78} &
\textbf{93.10$\pm$0.28} & 88.93 \\

& walker2d &
59.74$\pm$5.69 & 70.03$\pm$18.38 & 85.17$\pm$6.85 &
\textbf{108.52$\pm$0.71} & \textbf{108.14$\pm$0.19} &
\textbf{108.95$\pm$0.61} & 106.93 \\

& ant &
87.42$\pm$6.38 & 104.68$\pm$5.21 & 117.51$\pm$6.84 &
92.41$\pm$2.62 & 124.24$\pm$4.80 &
\textbf{137.89$\pm$0.45} & 130.75 \\ \hline

medium+few-expert & hopper &
50.74$\pm$2.92 & 52.02$\pm$2.64 & 56.79$\pm$11.55 &
38.07$\pm$11.56 & \textbf{104.89$\pm$5.22} &
97.24$\pm$2.22 & 111.34 \\

& halfcheetah &
41.49$\pm$1.28 & 41.60$\pm$2.99 & 46.33$\pm$7.44 &
66.07$\pm$8.52 & 53.44$\pm$26.85  &
\textbf{87.01$\pm$0.87} & 88.83 \\

& walker2d &
61.57$\pm$6.92 & 73.72$\pm$4.57 & 80.77$\pm$3.14 &
70.19$\pm$2.51 & \textbf{108.10$\pm$0.25} &
\textbf{108.02$\pm$0.76} & 106.93 \\

& ant &
86.13$\pm$7.24 & 89.11$\pm$1.53 & 101.26$\pm$8.73 &
92.55$\pm$5.60 & 119.69$\pm$2.22 &
\textbf{130.23$\pm$1.36} & 130.75 \\ \hline

cloned+expert & pen &
30.98$\pm$46.56 & 22.63$\pm$6.35 & 1.49$\pm$3.61 &
61.05$\pm$10.86 & \textbf{72.53$\pm$12.67} &
63.15$\pm$5.11 & 167.18 \\

& door &
0.27$\pm$1.84 & -0.08$\pm$0.08 & 1.17$\pm$2.16 &
\textbf{100.90$\pm$2.48} & 89.05$\pm$3.98 &
11.20$\pm$7.82 & 103.95 \\

& hammer &
0.59$\pm$0.19 & 0.48$\pm$0.46 & 6.00$\pm$5.44 &
54.29$\pm$23.25 & \textbf{68.46$\pm$17.79} &
37.84$\pm$8.05 & 125.72 \\ \hline

human+expert & pen &
73.84$\pm$18.47 & 47.64$\pm$14.83 & 27.71$\pm$6.32 &
\textbf{99.33$\pm$10.13} & 91.88$\pm$7.01 &
90.30$\pm$5.36 & 167.18 \\

& door &
47.21$\pm$12.72 & 1.50$\pm$1.06 & 0.06$\pm$0.03 &
96.45$\pm$4.55 & \textbf{100.16$\pm$6.84}  &
\textbf{101.43$\pm$3.73} & 104.73 \\

& hammer &
42.33$\pm$20.70 & 0.34$\pm$0.23 & 29.26$\pm$34.09 &
91.24$\pm$13.50 & 88.83$\pm$28.02 &
\textbf{109.63$\pm$9.16} & 125.72 \\ \hline

kitchen & partial & 0.0$\pm$0.0
 & 35.0$\pm$4.08 & 12.33$\pm$5.38 &
25.33$\pm$25.03 & \textbf{58.33$\pm$7.64} &
\textbf{57.25$\pm$6.81} & 75.00 \\

& mixed & 0.0$\pm$0.0
 & \textbf{48.33$\pm$4.24} & 7.50$\pm$4.16 &
21.00$\pm$6.06 & \textbf{46.67$\pm$1.44} &
\textbf{45.92$\pm$2.93} & 75.00 \\ \hline
\end{tabular}
}
\vspace{-1em}
\label{tab:main-result}
\end{table}

\noindent \textbf{Performance on MuJoCo Locomotion Benchmarks.} 
As shown in Table~\ref{tab:main-result}, TGE consistently matches or outperforms all baselines across environments. In \texttt{random+expert} and \texttt{medium+expert}, where the offline buffer contains relatively sufficient expert-like coverage, TGE remains highly competitive and achieves strong performance. The performance gains are most pronounced in \texttt{random+few-expert} and \texttt{medium+few-expert}, where expert trajectories are scarce. The DICE-based methods (SMODICE, DILO, PW-DICE) degrade substantially, 
while IOSTOM remains competitive on most tasks but underperforms on \texttt{halfcheetah} and \texttt{ant}.
Notably, TGE occasionally surpasses the provided expert demonstrations, as the single observation-only expert trajectory may not represent the best behavior in the dataset. By measuring trajectory-level proximity in the diffusion embedding space, TGE can recover stronger expert-like behaviors present in the offline buffer.

\noindent \textbf{Performance on Manipulation Benchmarks.}
On the high-dimensional Adroit, TGE performs strongly under \texttt{human+expert},
achieving the highest mean scores on \texttt{door} and
\texttt{hammer}, while remaining competitive on \texttt{pen}.
However, it underperforms leading occupancy-matching baselines
under \texttt{cloned+expert}, particularly on \texttt{door}.
One possible explanation is that latent similarity may reflect
task phase more strongly than task success. In stage-dependent
manipulation tasks, a segment may be closer to a failed segment
from the same phase than to a successful expert segment from
a different phase. Consequently, proximity in the embedding
space may not reliably indicate task-relevant progress.
On Franka Kitchen, TGE remains competitive with strong baselines
on both \texttt{partial} and \texttt{mixed}, demonstrating
its applicability beyond single-task manipulation.

\subsubsection{LfO from severe support mismatch}
We isolate the setting where the suboptimal dataset $\mathcal{D}_\mu$ 
contains \emph{no} expert trajectories at all, providing the cleanest 
empirical test of the support-mismatch analysis in Section~\ref{sec:analysis}. Two settings are considered: 
\textbf{medium}, where $\mathcal{D}_\mu$ consists of 1M transitions from a 
medium-quality policy, and \textbf{medium+random}, where $\mathcal{D}_\mu$ 
contains 1M equally mixed medium and random data. These settings are particularly 
challenging because they push the data distribution further from the expert support.

TGE+ReBRAC maintains strong performance across all tasks, while DICE-based baselines degrade sharply or collapse entirely as the data distribution shifts further from the expert. The results confirm the prediction of 
Section~\ref{sec:analysis}, that DICE-based baselines lose graded supervision as $\mathcal{D}_\mu$ drifts from the expert support, while TGE's distance-shaped reward in the diffusion latent space remains informative. Training-dynamics curves for the medium+random setting are provided in Figure~\ref{fig:training_dynamics_medrand} (Appendix~\ref{sec:training-dynamics} ).
\begin{table}[H]
\centering
\caption{\textbf{Results under severe support mismatch.} TGE+ReBRAC (highlighted 
in {\color{cyan}cyan}) maintains strong performance across all tasks, 
while several DICE-based baselines degrade substantially.}
\renewcommand{\arraystretch}{1.2}
\resizebox{\textwidth}{!}{
\begin{tabular}{|l|l|c|c|c|c|c|>
{\columncolor{cyan!10}}c|c|}
\hline
\textbf{Dataset} & \textbf{Env} & \textbf{Diffuser (BC)} & \textbf{SMODICE} & \textbf{PW-DICE} & \textbf{DILO} & \textbf{IOSTOM} & \textbf{TGE+ReBRAC} & \textbf{Expert} \\ \hline
medium & hopper &
46.33$\pm$1.86 & 56.18$\pm$1.16 & \textbf{65.82$\pm$11.39} &
48.68$\pm$17.85 & 59.50$\pm$7.45  &
\textbf{67.90$\pm$1.59} & 111.34 \\

& halfcheetah &
41.64$\pm$0.30 & 42.44$\pm$0.50 & 43.27$\pm$0.58 &
42.07$\pm$0.47 & 37.50$\pm$6.11 &
\textbf{48.00$\pm$0.30} & 88.83 \\

& walker2d &
65.62$\pm$4.89 & 71.28$\pm$7.21 & 75.64$\pm$3.58 &
69.74$\pm$1.13 & 72.35$\pm$5.65 &
\textbf{84.56$\pm$0.68} & 106.93 \\

& ant &
85.67$\pm$11.66 & 88.94$\pm$2.89 & 94.90$\pm$4.55 &
90.13$\pm$4.68 & 100.09$\pm$0.76 &
\textbf{124.12$\pm$1.79} & 130.75 \\ \hline

medium+random & hopper & 6.62$\pm$7.10
 & 54.86$\pm$2.49 & 51.77$\pm$16.16 &
58.71$\pm$3.86 & 48.20$\pm$11.74 &
\textbf{65.30$\pm$1.87} & 111.34 \\

& halfcheetah & 12.81$\pm$17.20
 & 16.49$\pm$2.32 & 3.17$\pm$1.69 & 2.25$\pm$0.01
 & 2.24$\pm$0.00 &
\textbf{44.22$\pm$0.14}& 88.83 \\

& walker2d &
0.96$\pm$2.06 & \textbf{78.93$\pm$6.07} & 41.46$\pm$38.23 & 46.52$\pm$14.22
 & 45.09$\pm$40.05  &
\textbf{79.12$\pm$1.21} & 106.93 \\

& ant &
 16.77$\pm$5.72 & 89.88$\pm$1.64 & 87.02$\pm$1.72 & 98.70$\pm$1.61
 &  92.86$\pm$8.14 &
\textbf{106.28$\pm$0.55} & 130.75 \\ \hline
\end{tabular}
}
\vspace{-1em}
\label{tab:mismatched-result}
\end{table}

\subsubsection{Comparison of reward constructions}
\label{app:seabo}

We compare five reward constructions using ReBRAC as the
downstream policy learner: a uniform reward, a distance-based
reward on raw observations, a reward based on VAE trajectory
embeddings, SEABO~\citep{lyu2024seabo}, and TGE. Both the VAE and TGE encoders use
trajectory windows of length $H=32$. This comparison evaluates
alternative reward constructions under a shared policy-learning
backbone.

\begin{table}[H]
\centering
\caption{Comparison of reward constructions with ReBRAC.
We report mean and standard deviation of normalized scores.
The highest mean in each setting is highlighted in bold.}
\label{tab:reward_comparison}
\small
\setlength{\tabcolsep}{3pt}
\begin{tabular}{lccccc}
\toprule
Environment & Uniform & Raw state & VAE & SEABO & TGE \\
\midrule
\multicolumn{6}{l}{\textit{Medium, without expert trajectories}} \\
hopper
& $59.97\pm1.51$ & $61.28\pm1.21$ & $60.37\pm2.53$
& $65.18\pm3.49$ & $\mathbf{67.90\pm1.59}$ \\
halfcheetah
& $42.80\pm0.25$ & $44.24\pm0.19$ & $46.39\pm0.12$
& $\mathbf{48.21\pm0.94}$ & $\mathbf{48.00\pm0.30}$ \\
walker2d
& $76.40\pm1.11$ & $80.20\pm0.45$ & $81.23\pm1.93$
& $75.61\pm1.76$ & $\mathbf{84.56\pm0.68}$ \\
ant
& $92.70\pm3.23$ & $114.47\pm0.44$ & $115.44\pm0.04$
& $108.17\pm1.54$ & $\mathbf{124.12\pm1.79}$ \\
\midrule
\multicolumn{6}{l}{\textit{Random + 30 unlabeled expert trajectories}} \\
hopper
& $35.37\pm26.52$ & $80.77\pm11.74$
& $\mathbf{98.63\pm20.13}$ & $82.22\pm21.46$
& $88.86\pm12.28$ \\
halfcheetah
& $2.25\pm0.01$ & $82.84\pm2.29$ & $\mathbf{87.65\pm1.91}$
& $67.12\pm4.91$ & $\mathbf{89.70\pm1.63}$ \\
walker2d
& $68.23\pm57.74$ & $101.24\pm11.36$ & $\mathbf{103.93\pm3.66}$
& $104.43\pm4.26$ & $\mathbf{105.48\pm2.52}$ \\
ant
& $37.75\pm7.05$ & $108.34\pm5.50$ & $\mathbf{111.15\pm7.85}$
& $96.65\pm12.66$ & $\mathbf{111.46\pm4.14}$ \\
\midrule
\multicolumn{6}{l}{\textit{Medium + 30 unlabeled expert trajectories}} \\
hopper
& $59.32\pm2.57$ & $81.64\pm8.24$
& $\mathbf{98.63\pm7.84}$ & $86.84\pm8.68$
& $\mathbf{97.24\pm2.22}$ \\
halfcheetah
& $82.84\pm2.29$ & $44.41\pm0.17$ & $89.91\pm1.76$
& $87.42\pm3.63$ & $\mathbf{91.83\pm0.80}$ \\
walker2d
& $77.84\pm1.26$ & $80.40\pm0.91$ & $\mathbf{107.29\pm1.86}$
& $77.81\pm1.86$ & $\mathbf{108.02\pm0.76}$ \\
ant
& $93.05\pm1.19$ & $119.11\pm1.31$ & $117.48\pm5.44$
& $115.30\pm1.91$ & $\mathbf{130.23\pm1.36}$ \\
\bottomrule
\end{tabular}
\end{table}

As shown in Table~\ref{tab:reward_comparison}, TGE achieves best or competitive performance in 11 of 12 settings and the highest average score across the evaluated settings. Its gains over the uniform reward indicate the value of expert-informed reward construction. Compared with raw-state distances and SEABO, TGE generally provides stronger performance, while the VAE baseline also performs well and achieves higher mean scores on two Hopper settings. These results support the utility of learned trajectory representations, with diffusion embeddings providing the strongest overall performance in this comparison.

\subsubsection{LfO from mismatched dynamics}
Beyond support mismatch in the data distribution, we evaluate TGE under dynamics mismatch, where the expert and the offline data are generated by agents with different physical embodiments. Following SMODICE~\citep{ma2022versatile}, we construct two settings: \textbf{HalfCheetah-Short}, where the expert operates with a halved torso, and \textbf{Ant-Disabled}, where the expert operates with two disabled legs. In both cases, $\mathcal{D}_E$ contains expert state 
trajectories from the modified embodiment, while $\mathcal{D}_\mu$ consists of suboptimal trajectories from the standard embodiment. The agent must imitate expert-like state visitation despite being unable 
to reproduce the expert's exact dynamics.

\begin{figure}[h]
\centering
\vspace{-0.5em}
\includegraphics[width=0.8\linewidth]{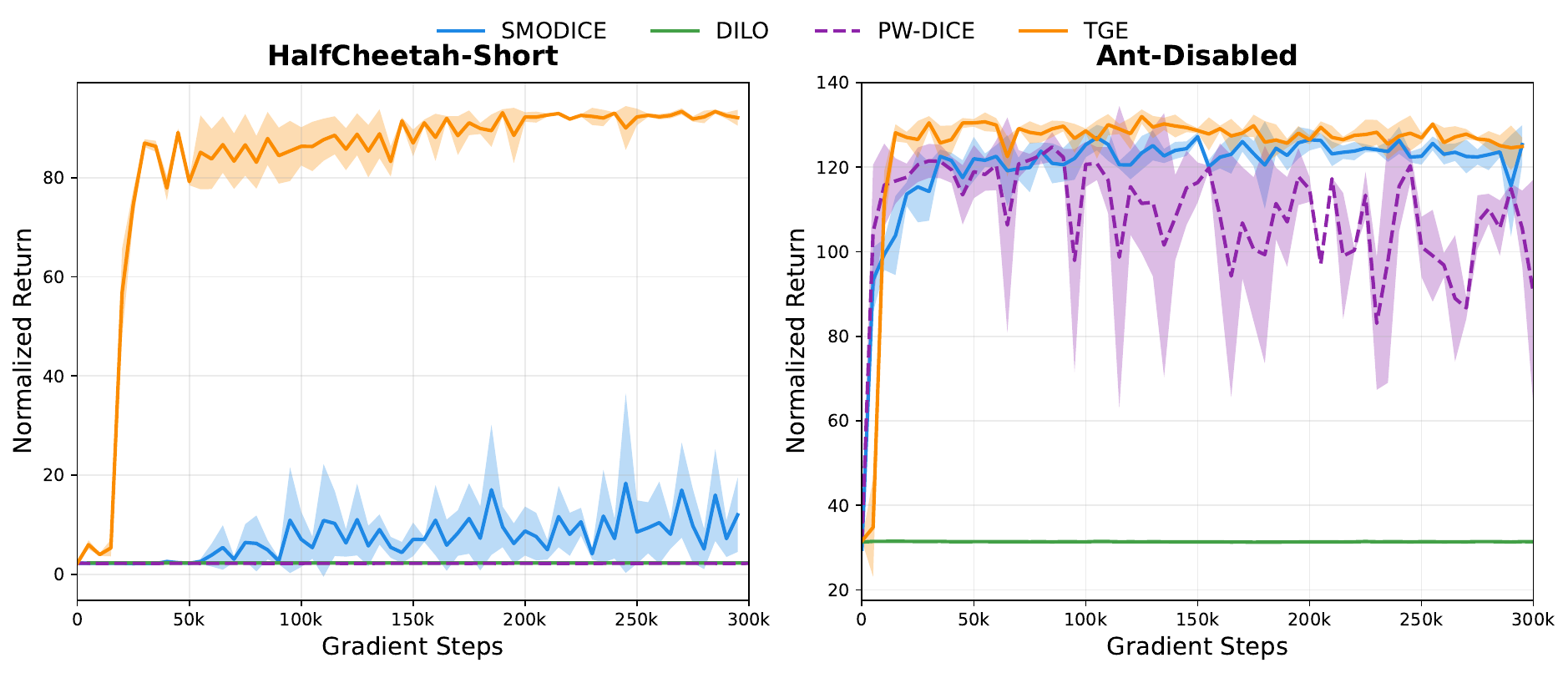}
\caption{\textbf{Training dynamics under mismatched dynamics.}\protect\footnotemark{} TGE achieves stable and rapid convergence to expert-level performance on both HalfCheetah-Short and Ant-Disabled, while DICE-based baselines either fail to converge or remain substantially below expert performance.}
    \label{fig:mismatched_dynamics}
\end{figure}
\footnotetext{We omit IOSTOM from this experiment because its released implementation does not support the
mismatched-dynamics setting, and our adaptation did not yield stable training despite hyperparameter tuning.}
As shown in Figure~\ref{fig:mismatched_dynamics}, TGE rapidly converges to near-expert performance on both tasks, while SMODICE plateaus well below expert performance on HalfCheetah-Short and DILO fails to learn a functional policy on either task. We attribute this robustness to the trajectory-level latent geometry. Even when the offline data is generated by a different embodiment, the diffusion encoder captures temporal dynamics that remain semantically meaningful relative to the expert state distribution, allowing a graded reward signal where occupancy-matching objectives become ill-conditioned.

\section{Conclusion}

In this work, we introduced Trajectory-level Generative Embeddings (TGE), a novel framework for offline imitation learning from observations that addresses the structural limitations of distribution-matching approaches. By shifting the paradigm from density ratio estimation to generative representation learning, 
TGE reduces reliance on expert-behavior support overlap that hampers prior methods. Motivated by the particle-based cross-entropy estimation, we demonstrated that encoding trajectories into the latent space of a temporal diffusion model allows for the construction of a dense, informative reward signal even when the support mismatch between offline data and the expert is severe.

These findings suggest that generative models can serve as powerful tools for structuring the state space in RL, going beyond their traditional roles in planning and policies. Future work could explore structured trajectory alignment to better distinguish task progress from phase similarity
in stage-dependent manipulation, as well as extending the framework to online fine-tuning and imitation from cross-embodiment observations with video diffusion models.

\begin{ack}
We thank the anonymous reviewers for their helpful comments. This research was supported by
WZ’s startup funding provided by the School of Data and Information Science at UNC Chapel Hill. This research was also supported by the NVIDIA Academic Grant Program with the cloud GPU credits and cloud infrastructure.
\end{ack}
\bibliographystyle{plainnat}
\bibliography{references}
\newpage

\appendix


\section{Hyperparameters and implementation details}
\label{sec:hyperparam}

\subsection{Hyperparameters}
We list the hyperparameters used for the diffuser, including details of the U-Net architecture, as well as those for TGE reward estimation in Table~\ref{tab:tge-hyperparam}. 
We also report the hyperparameters for the downstream offline RL backbones,ReBRAC and IQL, in Tables~\ref{tab:hyperparams_rebrac} and~\ref{tab:iql-hyperparam}, respectively. 
For all baseline methods, we strictly follow the original implementations and hyperparameter settings reported in their respective papers.

\begin{table}[H]
    \centering
    \caption{\textbf{Hyperparameter settings for TGE.} The hyperparameters below are used to train the trajectory-level generative planner and estimate the rewards.}
    \label{tab:hyperparams_tge}
    \begin{tabular}{l|c}
        \toprule
        \textbf{Hyperparameter} & \textbf{Value} \\
        \midrule
        \multicolumn{2}{l}{\textit{Reward Estimation}} \\
        \hline
        
        Kernel type & Logarithmic \\
        Kernel temperature ($\sigma$) & 1.0 \\
        Number of nearest neighbors ($m$) & 10 \\
        
        \hline
        \multicolumn{2}{l}{\textit{Diffuser Architecture}} \\
        \hline
        Diffuser horizon ($H$) & 32 \\
        Channel multipliers & (1, 2, 4, 8) \\
        Diffusion steps & 20 \\
        Embedding dimension & 256 \\
        Learning rate & 2e-4 \\
        Batch size & 32\\
        EMA decay & 0.995 \\
        Scheduler & DDPMScheduler\\
        \bottomrule
    \end{tabular}
    \label{tab:tge-hyperparam}
\end{table}

\begin{table}[H]
    \centering
    \caption{\textbf{Hyperparameter Settings for ReBRAC.} For the actor BC coefficient, values $\{0.1, 0.4, 1.0\}$ are used for \{Adroit-Cloned, Mujoco, Adroit-Human\}, respectively.}

    \label{tab:hyperparams_rebrac}
    \begin{tabular}{l|c}
        \toprule
        \textbf{Hyperparameter} & \textbf{Value} \\
        \midrule
        \multicolumn{2}{l}{\textit{Optimizer \& Architecture}} \\
        \hline
        Optimizer & AdamW \\
        Actor Learning rate & $5e-4$ \\
        Critic Learning rate & $5e-4$ \\
        Weight decay & $1e-3$ \\
        Batch size & 4096 \\
        Hidden layers & 3 \\
        Hidden dimension & 256 \\
        Layer normalization & True \\
        Actor Dropout & 0 (Mujoco) / 0.1(Adroit) \\
        \midrule
        \multicolumn{2}{l}{\textit{Algorithm Specifics}} \\
        \hline
        Discount factor ($\gamma$) & 0.99 \\
        Target smoothing ($\tau_{\text{target}}$) & 0.005 \\
        Policy update frequency & 2 \\
        Policy noise ($\sigma$) & 0.2 \\
        Noise clip & 0.5 \\
        Actor BC coefficient ($\beta_{\text{actor}}$) & 0.1, 0.4, 1.0 \\
        Critic BC coefficient ($\beta_{\text{critic}}$) & 1.0 \\
        \bottomrule
    \end{tabular}
    \label{tab:rebrac-hyperparam}
\end{table}

\begin{table}[H]
    \centering
    \caption{\textbf{Hyperparameter Settings for IQL.}}
    \label{tab:hyperparams_iql}
    \begin{tabular}{l|c}
        \toprule
        \textbf{Hyperparameter} & \textbf{Value} \\
        \midrule
        \multicolumn{2}{l}{\textit{Optimizer}} \\
        \hline
        Optimizer & AdamW \\
        Actor Learning rate & $3e-4$ \\
        Critic Learning rate & $3e-4$ \\
        Value Function Learning rate & $3e-4$ \\
        Batch size & 1024 \\
        \midrule
        \multicolumn{2}{l}{\textit{Algorithm Specifics}} \\
        \hline
        Discount factor ($\gamma$) & 0.99 \\
        Target smoothing ($\tau_{\text{target}}$) & 0.005 \\
        Expectile & 0.8 \\
        Inverse Temperature ($\beta$) & 0.5\\
        Dropout rate & 0 (Mujoco) / 0.1(Adroit) \\
        \bottomrule
    \end{tabular}
    \label{tab:iql-hyperparam}
\end{table}

\subsection{Implementation details}
\label{sec:impl}

We implement the TGE reward as the logarithmic distance to the nearest expert embeddings, consistent with the entropy minimization objective derived in Section~\ref{sec:motivation}. Since $\mathcal{D}_E$ contains only state observations, we mask the action dimensions of all trajectory segments by setting them to zero before encoding, applying the same 
masking to both expert and suboptimal segments for consistency. All embeddings are normalized to the unit hypersphere ($z \leftarrow z / \lVert z \rVert_2$). Distances are computed using Euclidean distance on the sphere, which is monotonic with cosine distance \mbox{($\lVert z_i-z_j\rVert_2=\sqrt{2-2\cos\theta}$)}.

For each offline transition, we retrieve its $m$ nearest expert neighbors in the embedding space (with $m{=}10$ by default) and compute the reward using a logarithmic kernel:
$r = -\frac{1}{m}\sum_{k=1}^m \log(1 + d_k / \sigma)$,
where $d_k$ is the distance to the $k$-th neighbor and $\sigma$ is a temperature scaling parameter. This reward signal is then normalized and used to label transitions in $\mathcal{D}_\mu$.

Finally, we train offline policies using IQL and ReBRAC as drop-in backbones, relying solely on $\mathcal{D}_\mu$ and the learned TGE reward, without any online interaction. Additionally, to ensure stable learning and balanced sampling across the diverse trajectories in the suboptimal dataset, we utilize the normalized TGE reward as a weighting factor for the behavioral cloning term in our ReBRAC backbone. We treat rewards as fixed confidence weights that scale the MSE penalty between the actor's actions and the dataset actions. This ensures that the policy regularizes more strongly toward expert-labeled regions (high reward) while maintaining flexibility in suboptimal regions, effectively preventing the policy from collapsing toward poor-quality data points.

\subsection{Computational overhead}
As noted in the experimental evaluation, TGE involves a multi-stage pre-processing pipeline before policy learning begins. We report wall-clock times on Hopper-Random+Expert using a single NVIDIA L40S GPU in Table~\ref{tab:runtime}.

\begin{table}[h]
\centering
\caption{Wall-clock time in seconds. Total runtime includes
300K policy updates for TGE and SEABO, and 500K updates for
IOSTOM and DILO, following their respective convergence settings.}
\label{tab:runtime}
\begin{tabular}{lrrrrr}
\toprule
Method & Pretraining & Annotation
       & \shortstack{RL\\300K}
       & \shortstack{RL\\500K}
       & Total \\
\midrule
TGE    & $\approx$3,600 & 42.9 & 448.7   & --      & $\approx$4,092 \\
SEABO  & --            & 30.5 & 444.9   & --      & 475.4 \\
IOSTOM & --            & --   & 413.2   & 680.7   & 680.7 \\
DILO   & --            & --   & 2,287.2 & 3,841.5 & 3,841.5 \\
\bottomrule
\end{tabular}
\end{table}

TGE's main computational overhead is diffusion encoder pretraining.
Including this stage, its total runtime exceeds that of SEABO
and IOSTOM and is comparable to DILO. However, for a fixed offline dataset, the pretrained encoder can be reused
across expert demonstrations, reward relabeling runs, and downstream
policy training. With a pretrained encoder, annotation and 300K
policy updates take only 491.6 seconds in this setting, which is lower than the runtimes of IOSTOM and DILO at their respective
convergence budgets.

\section{Details of the mixture strategy for the suboptimal dataset}
We provide a detailed breakdown of the data mixture strategy used to construct the suboptimal dataset $\mathcal{D}_\mu$. Following established protocols in offline imitation learning from observations introduced in~\citep{ma2022versatile, sikchi2025dual}, we construct $\mathcal{D}_\mu$ by combining a large corpus of suboptimal transitions with a limited number of expert trajectories. Table~\ref{tab:dataset_composition} summarizes the detailed composition for all suboptimal datasets.

\label{sec:mixture-strategy}
\begin{table}[h]
    \centering
    \caption{\textbf{Dataset Mixture Details.} We summarize the data mixture strategies for the suboptimal datasets $\mathcal{D}_\mu$ used across different experimental settings.}
    \label{tab:dataset_composition}
    \begin{tabular}{l|c c c}
        \toprule
        \textbf{Dataset Name} & \textbf{Suboptimal Data} & \textbf{Expert Trajectories} & \textbf{Episode Length} \\
        \midrule
        \multicolumn{4}{l}{\textit{Locomotion Tasks}} \\
        \hline
        Random + Expert & 1e6 transitions & 200 & 1000 \\
        Medium + Expert & 1e6 transitions & 200 & 1000 \\
        Random + Few-Expert & 1e6 transitions & 30 & 1000 \\
        Medium + Few-Expert & 1e6 transitions & 30 & 1000 \\
        Medium & 1e6 transitions & 0 & 1000 \\
        Medium + Random& 1e6 transitions & 0 & 1000 \\
        \midrule
        \multicolumn{4}{l}{\textit{Adroit Tasks}} \\
        \hline
        Pen (Cloned + Expert) & 5e5 transitions & 30 & 100 \\
        Pen (Human + Expert) & 5e3 transitions & 30 & 100 \\
        Door (Cloned + Expert) & 1e6 transitions & 30 & 200 \\
        Door (Human + Expert) & 6.7e3 transitions & 30 & 200 \\
        Hammer (Cloned + Expert) & 1e6 transitions & 30 & 200 \\
        Hammer (Human + Expert) & 1.1e4 transitions & 30 & 200 \\
        Kitchen-Partial & 1.37e5 transitions & 18 & 280 \\
        Kitchen-Mixed & 1.37e5 transitions & 18 & 280 \\
        \bottomrule
    \end{tabular}
\end{table}

\section{Additional ablation studies}
\label{sec:additional-ablation}
\subsection{Impact of horizons in the generative planner}
\label{sec:horizon_ablation}
\begin{figure}[H]
\centering
\includegraphics[width=0.32\linewidth]{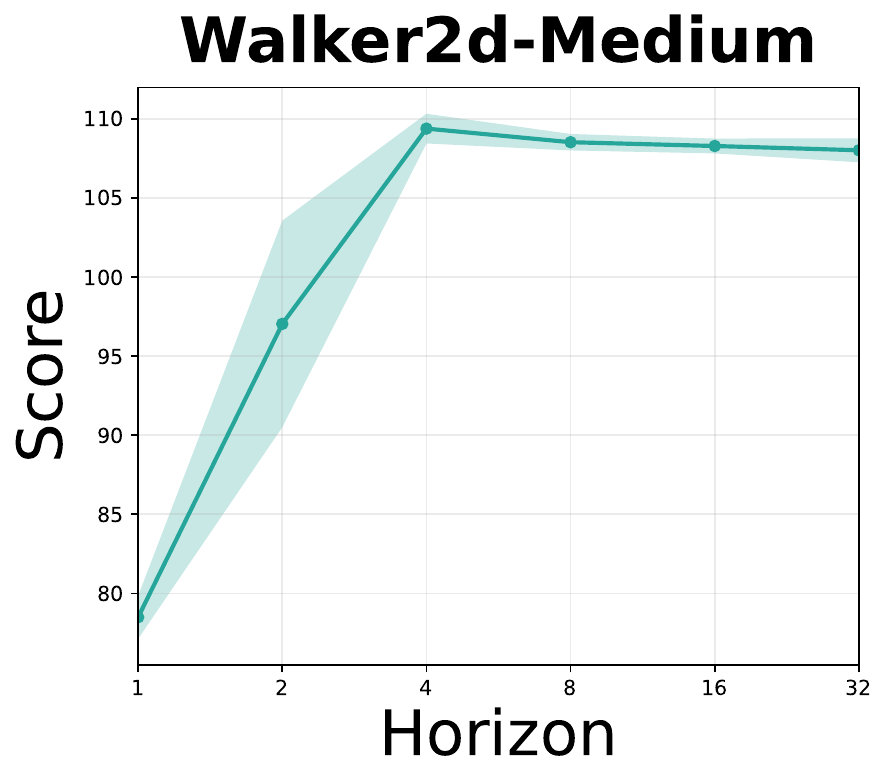}
\includegraphics[width=0.32\linewidth]{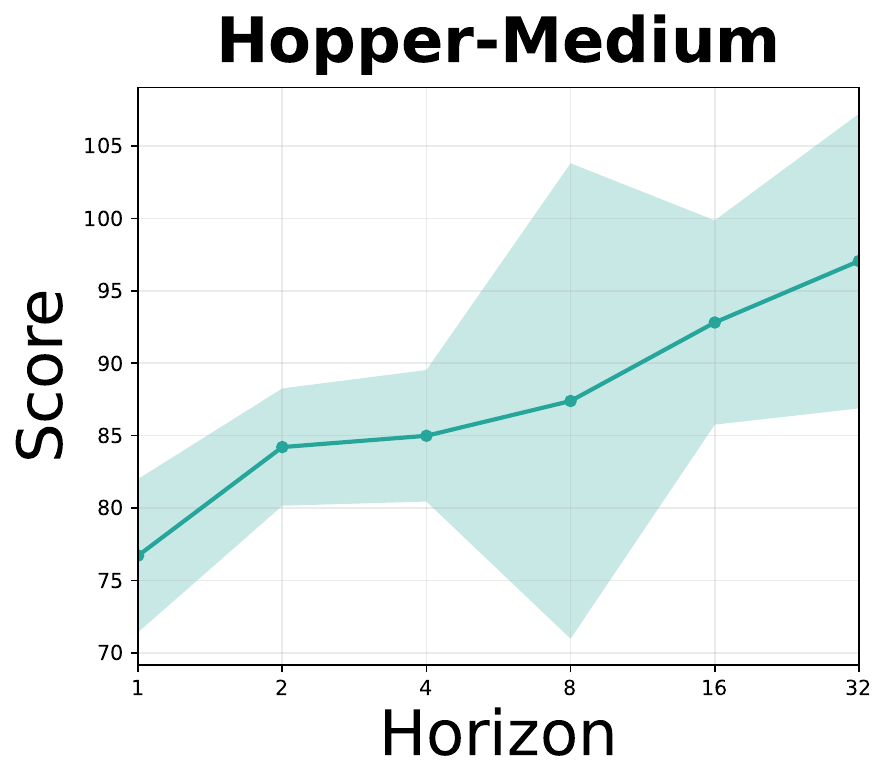}
\includegraphics[width=0.32\linewidth]{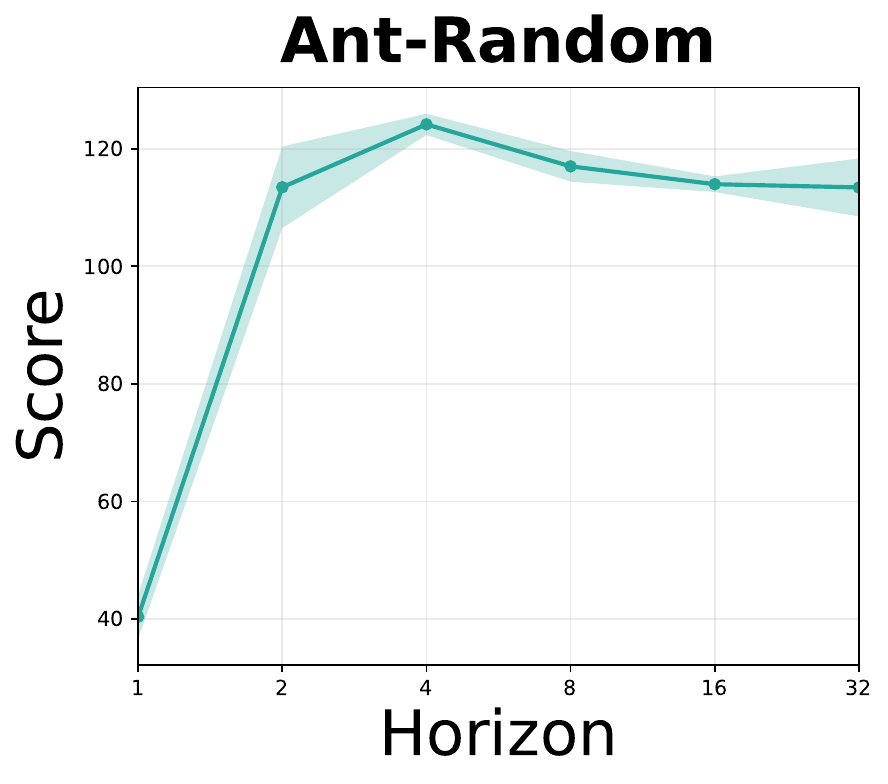}
\caption{\textbf{Ablation of the Temporal Horizon.} The results demonstrate how performance varies with the context horizon~$H$. We observe that larger~$H$ generally leads to improved performance, highlighting the importance of temporal context.}
\label{fig:horizon_ablation}
\end{figure}
We examine how the length of temporal context used by the generative planner influences reward estimation quality and the performance of the downstream offline RL policy. 
Specifically, we vary the horizon length~$H$ employed during planner training and embedding computation. 
As shown in Figure~\ref{fig:horizon_ablation}, using a single-step encoding ($H{=}1$) leads to a substantial degradation in the final policy performance, highlighting the importance of incorporating temporal context along the trajectory. 
Increasing~$H$ yields nearly consistent performance improvements across tasks when $H$ is small, and saturates after certain horizon lengths. 
This trend suggests that short-to-mid temporal windows are sufficient to capture essential dynamic information. While longer horizons maintain consistent performance without significant degradation, 
they 
may introduce unnecessary computation or over-smoothing effects. 
Based on these observations, we adopt a conservative default of $H{=}32$ for a unified and stable hyperparameter choice across all environments and configurations. Additional ablation studies on other aspects of our proposed method is deferred to Appendix~\ref{sec:additional-ablation}.

\subsection{Choice of kernels for reward estimation} 
\begin{figure}[H]
    \centering

    \begin{subfigure}[b]{0.66\linewidth}
        \centering
        \includegraphics[width=\linewidth]{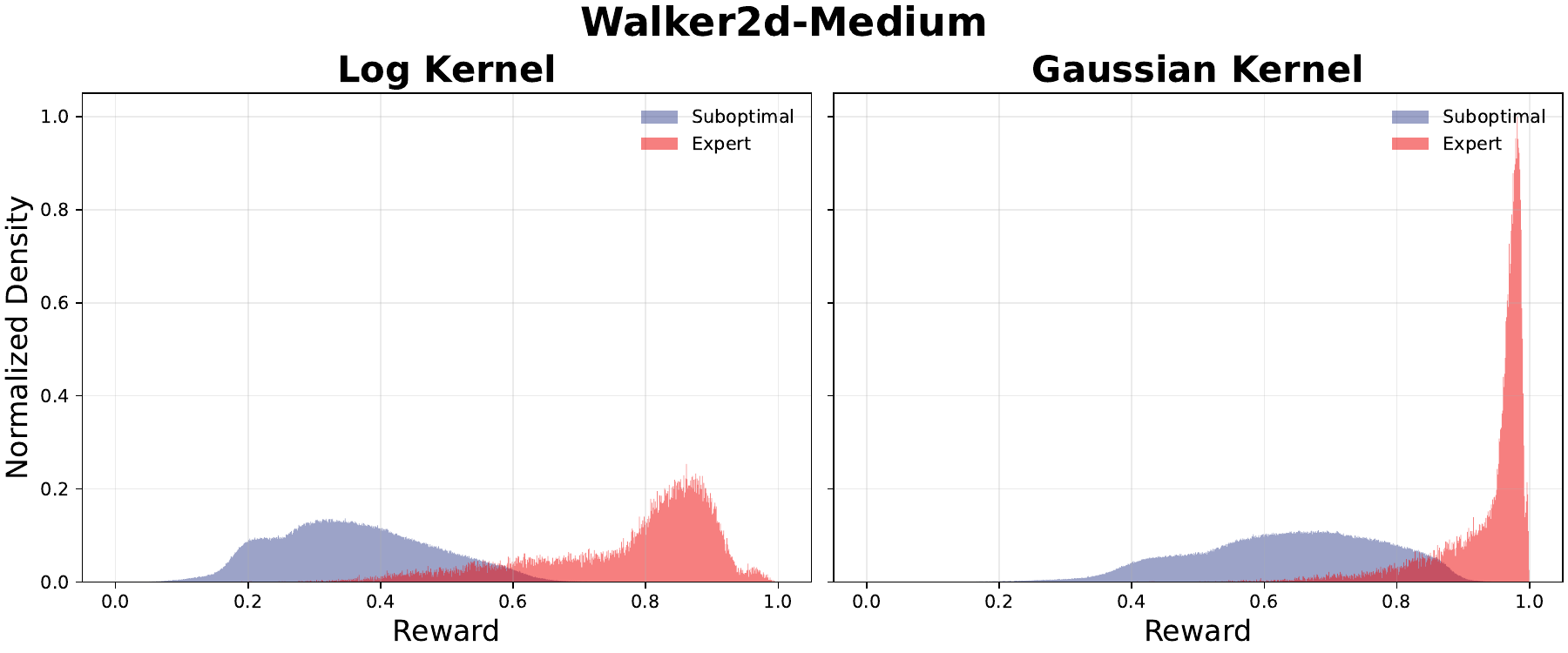}
    \end{subfigure}
    \hfill
    \begin{subfigure}[b]{0.32\linewidth}
        \centering
        \includegraphics[width=\linewidth]{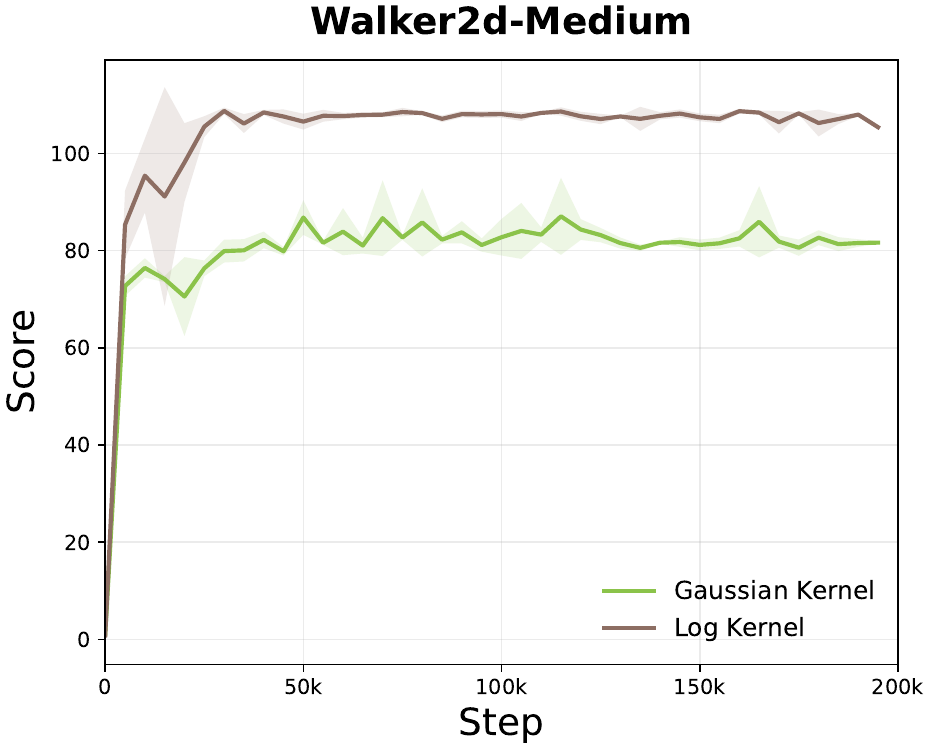}
    \end{subfigure}


    \begin{subfigure}[b]{0.66\linewidth}
        \centering
        \includegraphics[width=\linewidth]{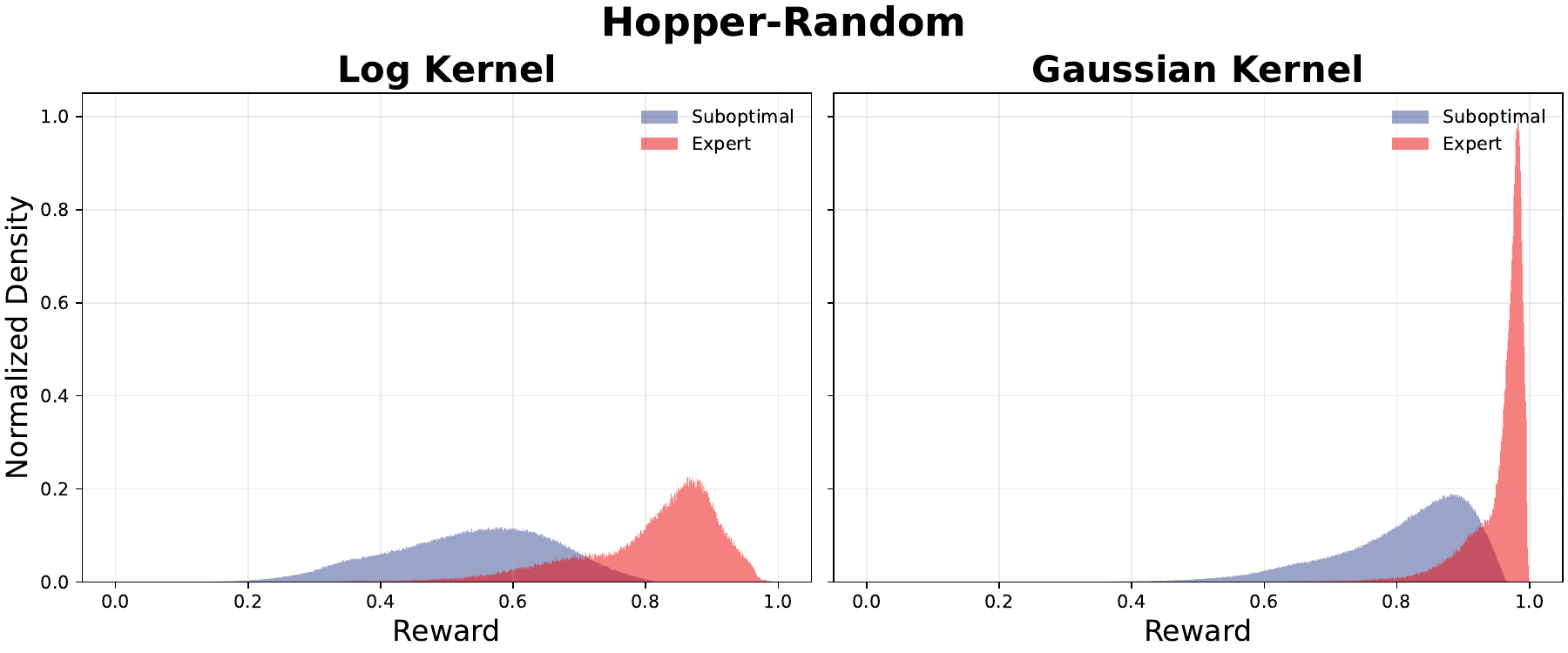}
        \caption{Reward Distribution Comparison}
    \end{subfigure}
    \hfill
    \begin{subfigure}[b]{0.32\linewidth}
        \centering
        \includegraphics[width=\linewidth]{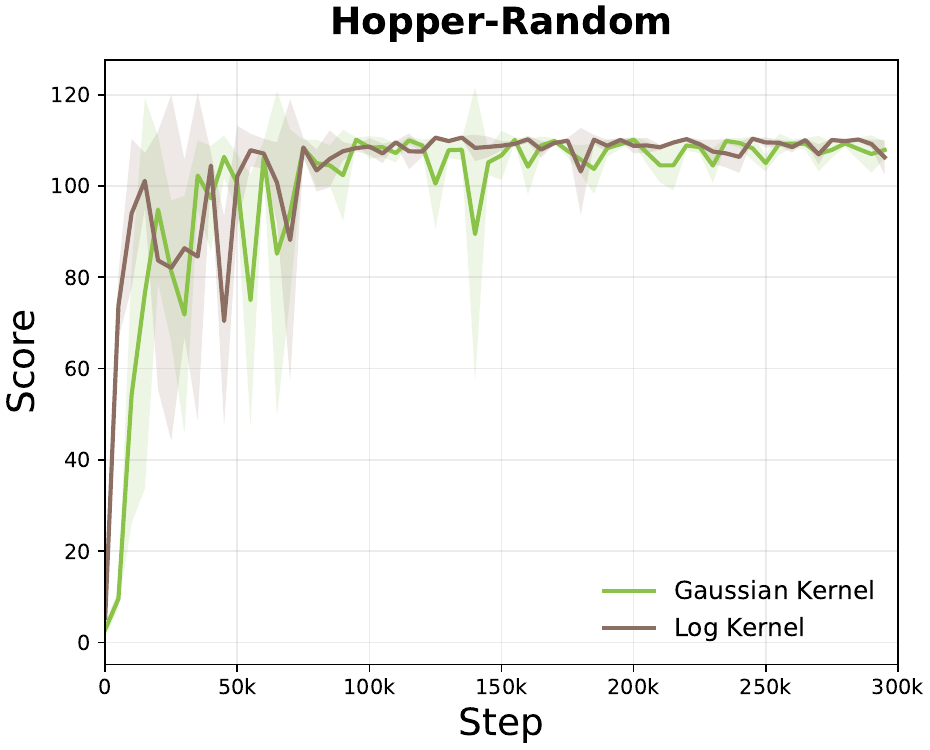}
        \caption{Performance}
    \end{subfigure}

    \caption{\textbf{Ablation of the Reward Signal Density.}
    (Top) Walker2d-Medium. (Bottom) Hopper-Random. 
    Distributions of synthetic rewards assigned to expert ({\color{red}red}) and suboptimal ({\color{blue}blue}) data.
    Compared to Gaussian-kernel-based reward estimation, the Logarithmic kernel preserves a heavier-tailed reward signal and is associated with improved performance and training stability.}

    \label{fig:ablation_kernel_main}
\end{figure}

We investigate the choice of kernel function $f(d)$ by analyzing the resulting reward distributions and their associated training dynamics. 
Figure~\ref{fig:ablation_kernel_main}(a) visualizes the density of synthetic rewards assigned to expert and suboptimal trajectories in the offline dataset under Logarithmic and Gaussian kernels, respectively.

As shown in Figure~\ref{fig:ablation_kernel_main}(a), the Gaussian kernel produces a reward distribution that is highly concentrated in narrow regions: expert rewards collapse near $1.0$, forming a sharp peak, while suboptimal trajectories cluster nearby. 
As a result, meaningful behavioral differences correspond to only minor reward variations, which limits the clarity of the learning signal. 
In contrast, the Logarithmic kernel (Figure~\ref{fig:ablation_kernel_main}(a, left)) yields a broader, heavy-tailed reward distribution. 
The slow decay better preserves the relative ordering of trajectories in the embedding space, improving reward separability. 
This enhanced distinguishability facilitates more accurate value estimation by the critic and leads to more stable and optimal learning curves, as shown in Figure~\ref{fig:ablation_kernel_main}(b).

\subsection{Learning from fragmented expert demonstrations}
\label{app:fragmented_expert}

We evaluate whether TGE requires complete expert trajectories
by constructing fragmented expert demonstrations. We randomly
extract non-overlapping contiguous clips of length $H=32$ from
multiple expert episodes, discard their original episode identities
and global temporal ordering, and treat each clip as an independent
fragment. Embeddings are computed within individual fragments,
without crossing fragment boundaries.

We compare three settings: \textbf{Complete}, which uses one
complete expert episode; \textbf{Fragment-1$\times$}, which uses
fragments containing the same total number of expert observations;
and \textbf{Fragment-0.25$\times$}, which retains only 25\% of that
observation budget.

\begin{table}[H]
\centering
\caption{Performance with complete and fragmented expert
demonstrations. Fragmented settings retain local continuity
while discarding global episode structure.}
\label{tab:fragmented_expert}
\begin{tabular}{lccc}
\toprule
Environment & Complete
& Fragment-1$\times$ & Fragment-0.25$\times$ \\
\midrule
hopper-medium
& $67.90 \pm 1.59$ & $67.80 \pm 3.75$ & $67.08 \pm 3.70$ \\
halfcheetah-medium
& $48.00 \pm 0.30$ & $45.11 \pm 0.18$ & $45.22 \pm 0.06$ \\
walker2d-medium
& $84.56 \pm 0.68$ & $83.67 \pm 0.10$ & $82.80 \pm 0.56$ \\
ant-medium
& $124.12 \pm 1.79$ & $118.45 \pm 0.14$ & $121.34 \pm 0.76$ \\
\bottomrule
\end{tabular}
\end{table}

As shown in Table~\ref{tab:fragmented_expert}, TGE maintains
broadly comparable performance with fragmented demonstrations,
although some degradation occurs on individual tasks.
Performance also remains close to the complete-episode baseline
when the expert observation budget is reduced to 25\%.
These results suggest that complete expert episodes and global
trajectory ordering are not necessary for TGE.

\subsection{Sensitivity to expert demonstrations}
\label{app:expert_sensitivity}

We examine sensitivity to the number of expert demonstrations
and the choice of the expert episode.

\paragraph{Number of expert episodes.}
We vary the number of expert episodes from 1 to 3, 5, and 10
while keeping the ReBRAC configuration fixed.
As shown in Table~\ref{tab:expert_count}, performance changes
little on the two evaluated tasks, suggesting that additional
expert episodes provide limited benefit in these settings.

\begin{table}[H]
\centering
\caption{Sensitivity to the number of expert episodes.}
\label{tab:expert_count}
\begin{tabular}{lcccc}
\toprule
Environment & 1 episode & 3 episodes & 5 episodes & 10 episodes \\
\midrule
hopper-medium
& $67.90 \pm 1.59$ & $67.78 \pm 2.36$
& $67.92 \pm 0.32$ & $68.17 \pm 3.35$ \\
walker2d-medium
& $84.56 \pm 0.68$ & $83.94 \pm 1.07$
& $84.10 \pm 0.66$ & $83.74 \pm 0.43$ \\
\bottomrule
\end{tabular}
\end{table}

\paragraph{Choice of expert episode.}
We evaluate TGE using five independently selected single expert
episodes. Table~\ref{tab:expert_selection} reports the score for
each selection and the sample standard deviation across the five
scores. Performance varies modestly across expert selections
in the evaluated locomotion tasks, indicating that the results
are not specific to a single expert episode.

\begin{table}[H]
\centering
\caption{Sensitivity to the choice of a single expert episode.
The final column reports the sample standard deviation across
the five episode selections, using $\mathrm{ddof}=1$.}
\label{tab:expert_selection}
\begin{tabular}{lrrrrrr}
\toprule
Environment & Episode 0 & Episode 1 & Episode 2
& Episode 3 & Episode 4 & Std. \\
\midrule
hopper-medium
& 67.90 & 69.92 & 64.20 & 70.60 & 64.53 & 2.97 \\
halfcheetah-medium
& 48.00 & 47.26 & 47.25 & 47.27 & 47.86 & 0.37 \\
walker2d-medium
& 84.56 & 84.01 & 84.48 & 84.37 & 83.31 & 0.51 \\
ant-medium
& 124.12 & 123.19 & 121.85 & 123.82 & 123.14 & 0.87 \\
\bottomrule
\end{tabular}
\end{table}

\subsection{Adaptive BC weighting in ReBRAC}
\label{app:bc_weighting}

In our ReBRAC backbone, we weight the behavioral cloning penalty by the normalized TGE reward, scaling the MSE between the actor's actions and the dataset actions according to the expert-likeness of each transition. This serves as a stabilization mechanism analogous to advantage-weighted behavioral cloning. As the surrogate rewards are inferred rather 
than ground truth, the weighted BC term prevents the policy from deviating too aggressively based on potentially noisy reward estimates in low-quality regions of $\mathcal{D}_\mu$. Table~\ref{tab:bc_weighting} shows that the weighting improves both training stability and final performance.

\begin{table}[H]
\centering
\caption{
Representative results with and without adaptive BC weighting.}
\label{tab:bc_weighting}
\begin{tabular}{lcc}
\toprule
Environment & w/o weighting & w/ weighting (default) \\
\midrule
walker2d-medium
& 80.95 $\pm$ 1.36 & 84.56 $\pm$ 0.68 \\
hopper-medium
& 63.14 $\pm$ 2.46 & 67.90 $\pm$ 1.59 \\
halfcheetah-medium
& 44.78 $\pm$ 0.08 & \textbf{48.00 $\pm$ 0.30} \\
halfcheetah-random+few-expert
& 67.00 $\pm$ 8.78 & \textbf{89.70 $\pm$ 1.63} \\
walker2d-medium+few-expert
& 109.08 $\pm$ 0.59 & 108.02 $\pm$ 0.76 \\
hopper-random+few-expert
& 91.09 $\pm$ 9.24 & 88.86 $\pm$ 12.28 \\
door-human+expert
& 57.89 $\pm$ 5.55 & \textbf{101.43 $\pm$ 3.73} \\
hammer-human+expert
& 93.33 $\pm$ 7.65 & \textbf{109.63 $\pm$ 9.16} \\
\bottomrule
\end{tabular}
\end{table}

\section{Training dynamics visualizations}
\label{sec:training-dynamics}
\subsection{TGE with IQL and ReBRAC backbones}
\begin{figure}[h]
    \centering
    \begin{subfigure}{0.24\textwidth}
        \includegraphics[width=\linewidth]{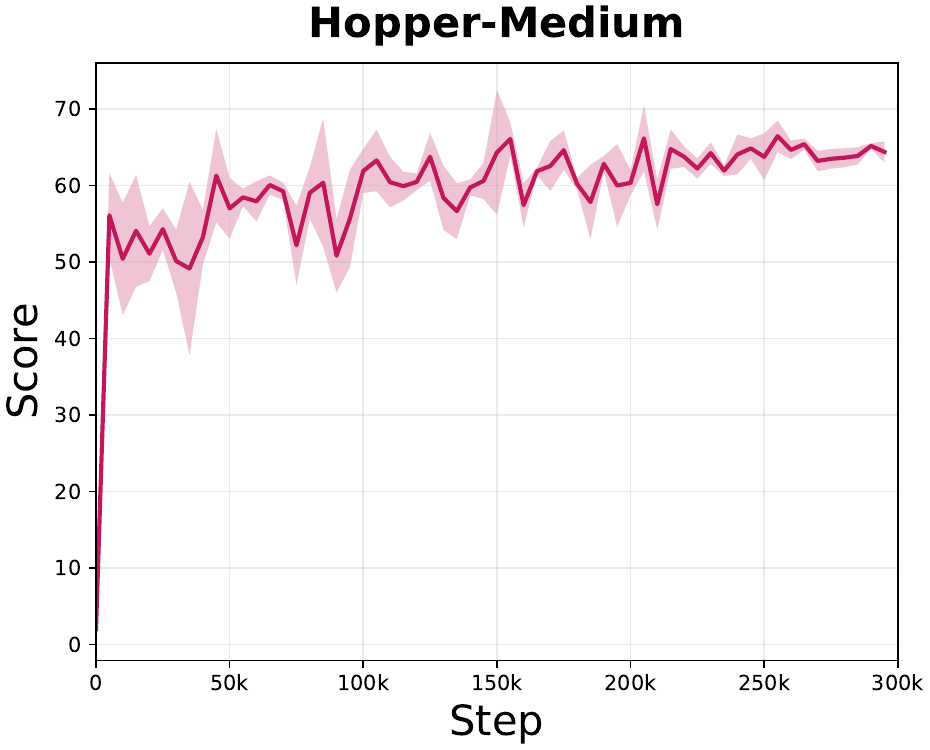}
        \caption{Hopper-Medium}
    \end{subfigure}
    \hfill
    \begin{subfigure}{0.24\textwidth}
        \includegraphics[width=\linewidth]{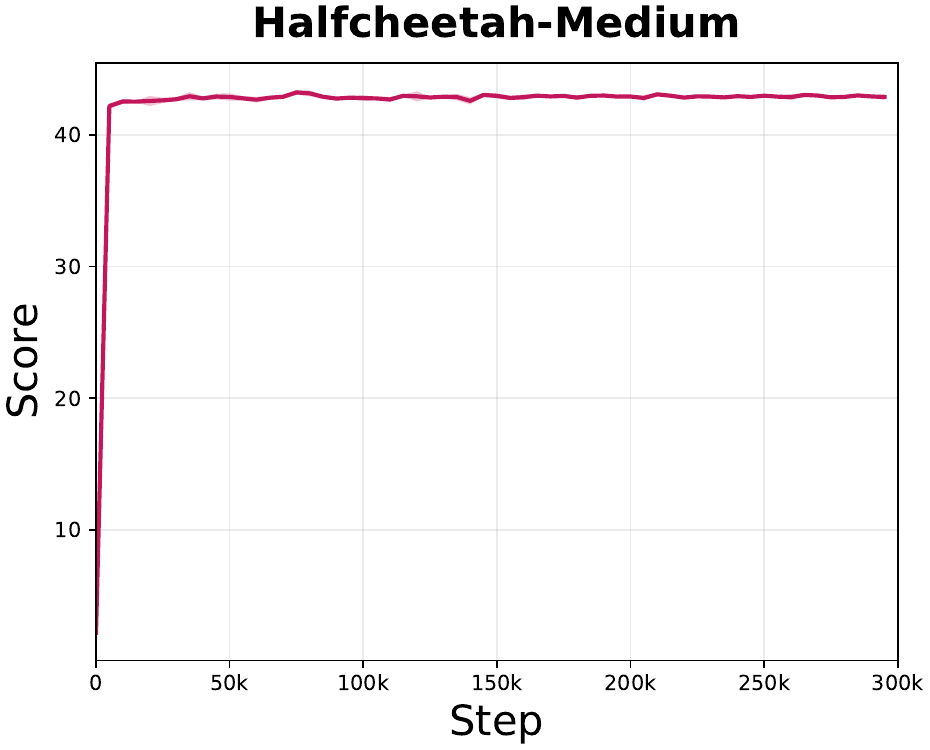}
        \caption{HalfCheetah-Medium}
    \end{subfigure}
    \hfill
    \begin{subfigure}{0.24\textwidth}
        \includegraphics[width=\linewidth]{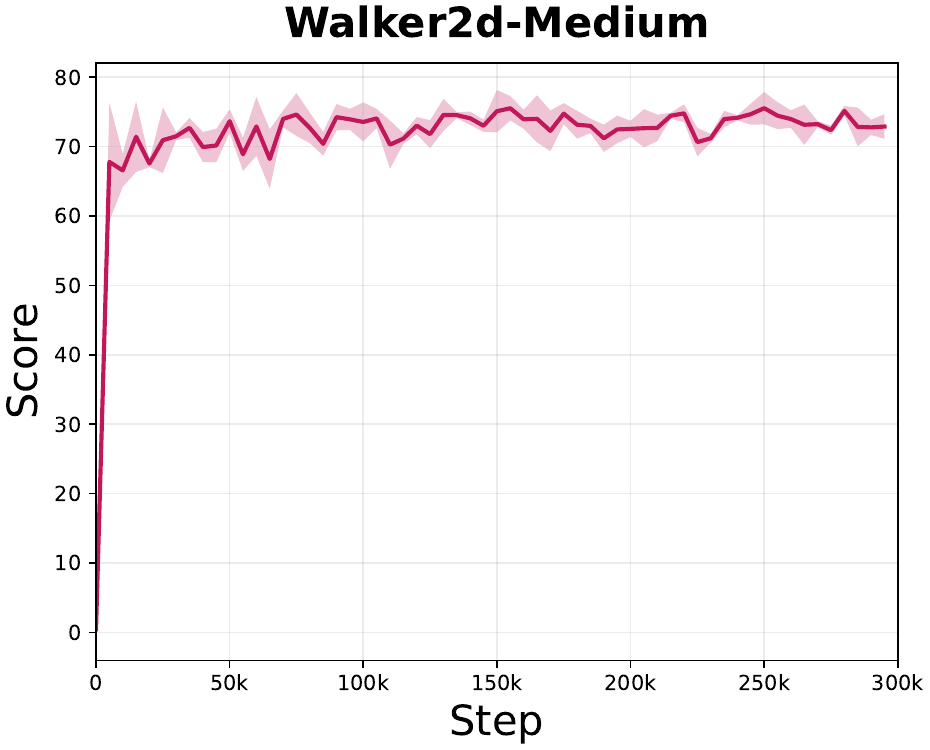}
        \caption{Walker2d-Medium}
    \end{subfigure}
    \hfill
    \begin{subfigure}{0.24\textwidth}
        \includegraphics[width=\linewidth]{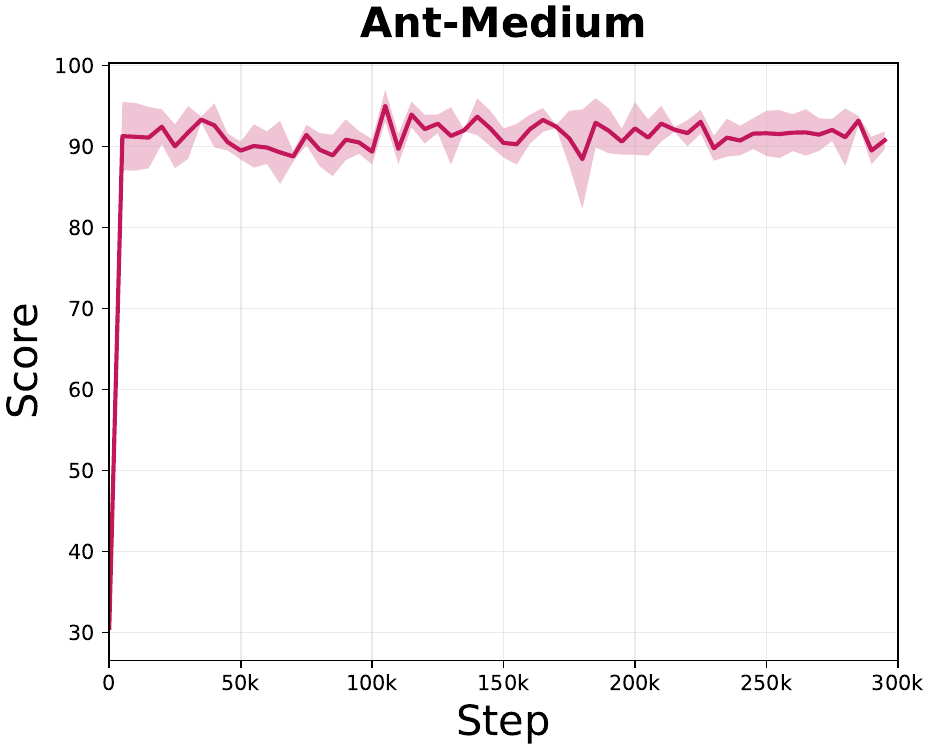}
        \caption{Ant-Medium}
    \end{subfigure}
    \hfill
    \begin{subfigure}{0.24\textwidth}
        \includegraphics[width=\linewidth]{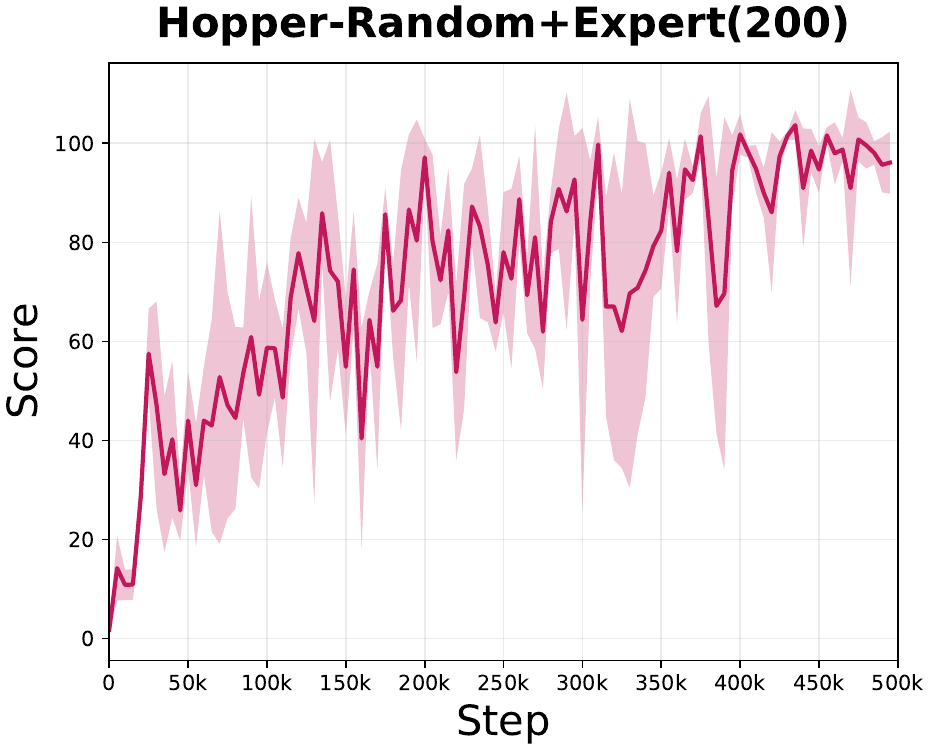}
        \caption{Hopper-Random}
    \end{subfigure}
    \hfill
    \begin{subfigure}{0.24\textwidth}
        \includegraphics[width=\linewidth]{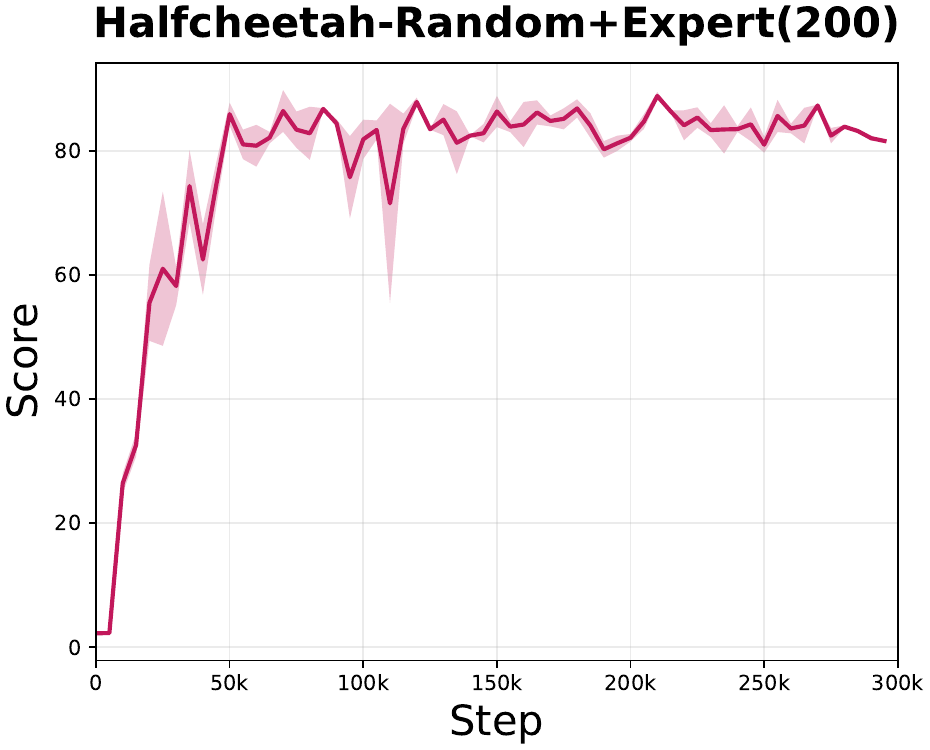}
        \caption{HalfCheetah-Random}
    \end{subfigure}
    \hfill
    \begin{subfigure}{0.24\textwidth}
        \includegraphics[width=\linewidth]{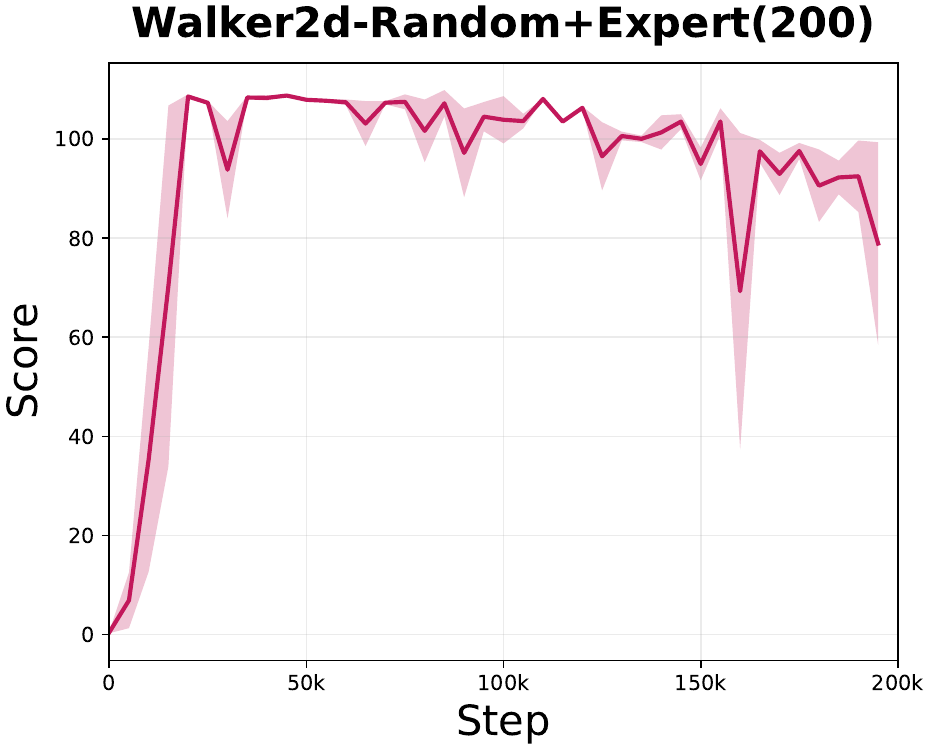}
        \caption{Walker2d-Random}
    \end{subfigure}
    \hfill
    \begin{subfigure}{0.24\textwidth}
        \includegraphics[width=\linewidth]{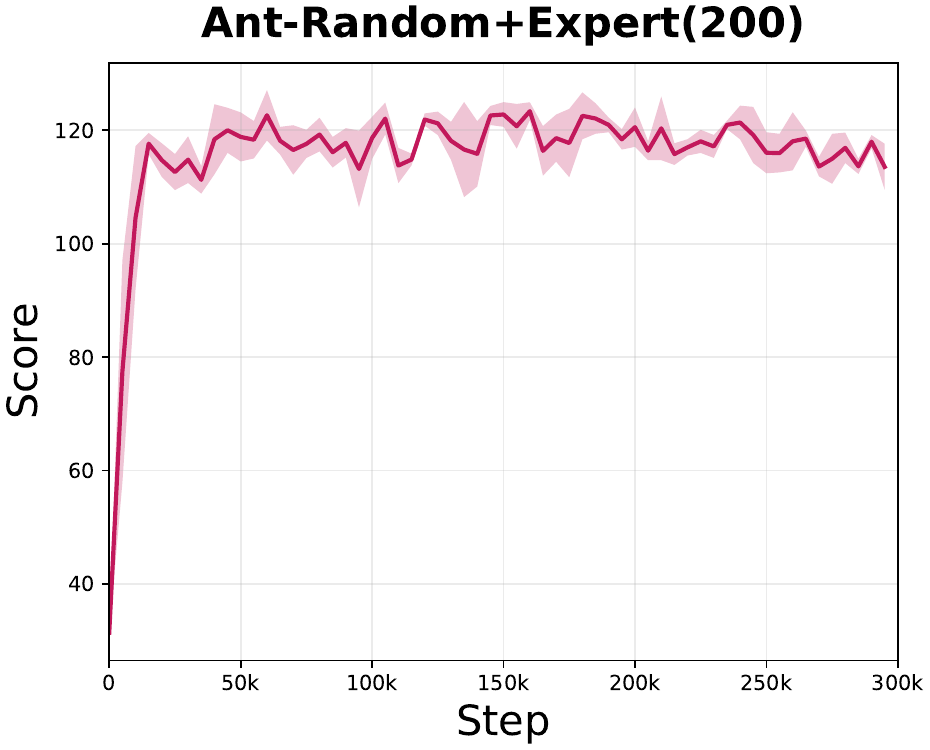}
        \caption{Ant-Random}
    \end{subfigure}
    \caption{\textbf{Training Dynamics of TGE + IQL.} The curves display the mean normalized score and standard deviation (shaded region). TGE combined with IQL shows stable policy improvement, confirming that our geometric reward signal enables robust learning across different offline RL methods.}
    \label{fig:iql_curves}
\end{figure}

\begin{figure}[h]
    \centering
    \begin{subfigure}{0.24\textwidth}
        \includegraphics[width=\linewidth]{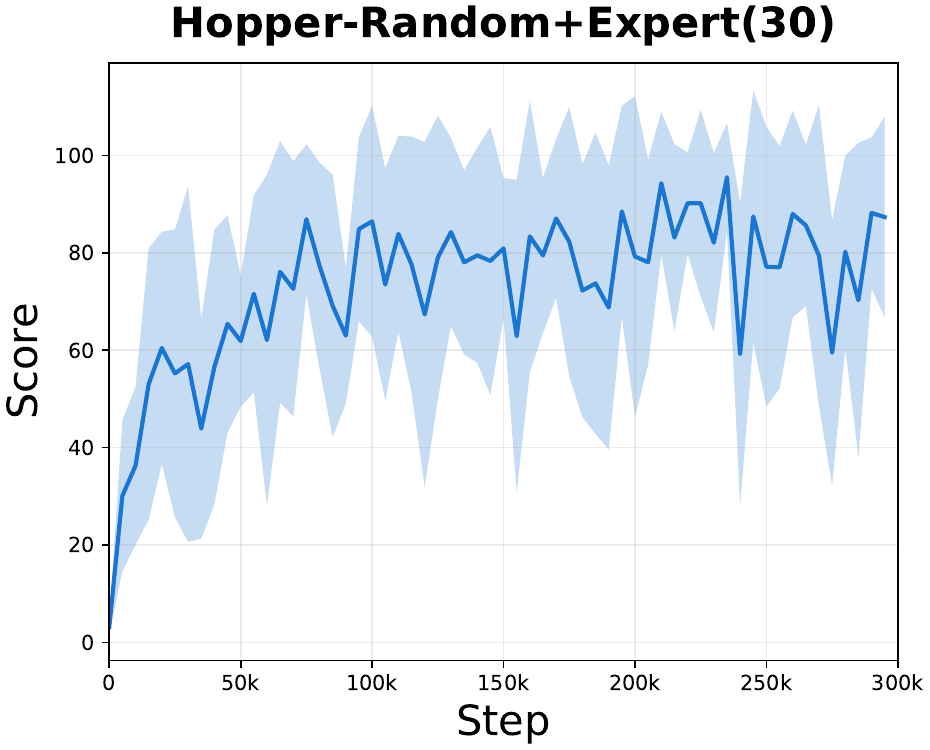}
        \caption{Hopper-Random}
    \end{subfigure}
    \hfill
    \begin{subfigure}{0.24\textwidth}
        \includegraphics[width=\linewidth]{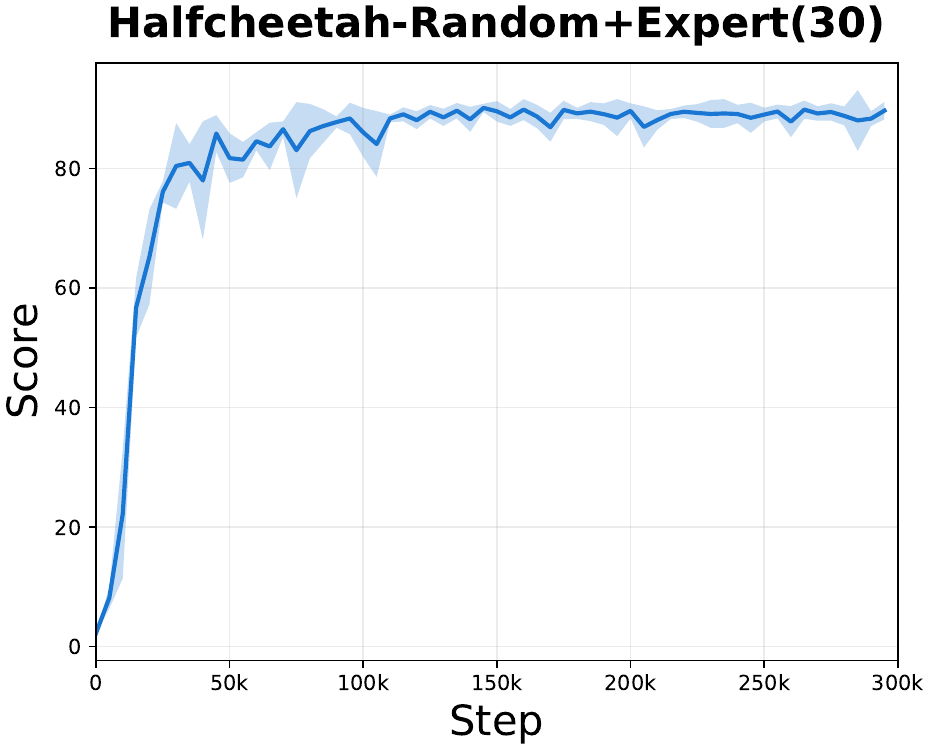}
        \caption{HalfCheetah-Random}
    \end{subfigure}
    \hfill
    \begin{subfigure}{0.24\textwidth}
        \includegraphics[width=\linewidth]{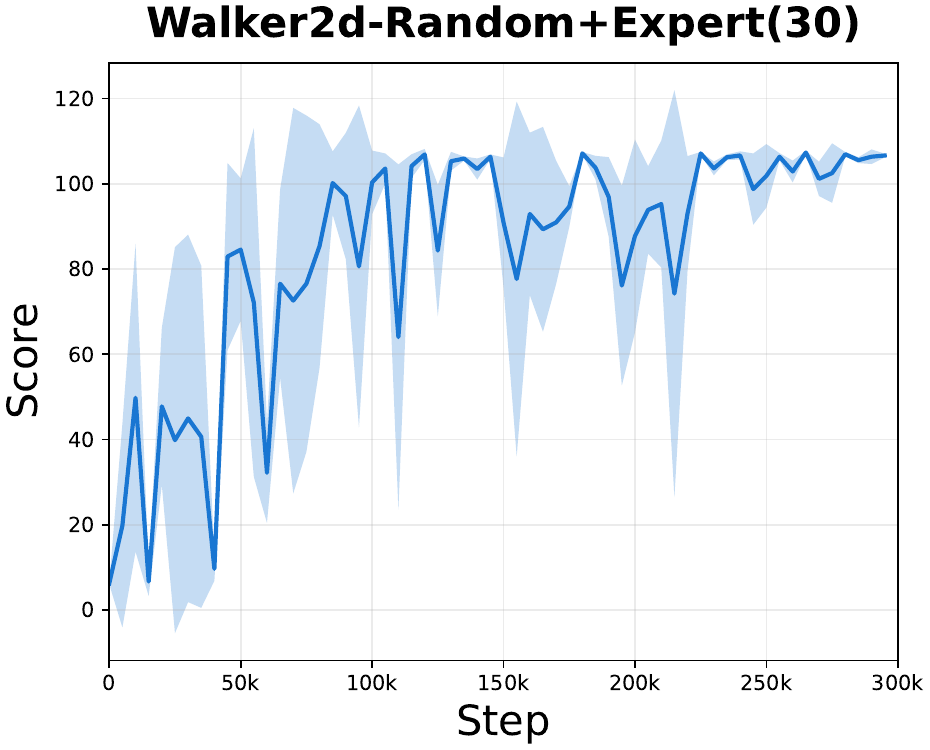}
        \caption{Walker2d-Random}
    \end{subfigure}
    \hfill
    \begin{subfigure}{0.24\textwidth}
        \includegraphics[width=\linewidth]{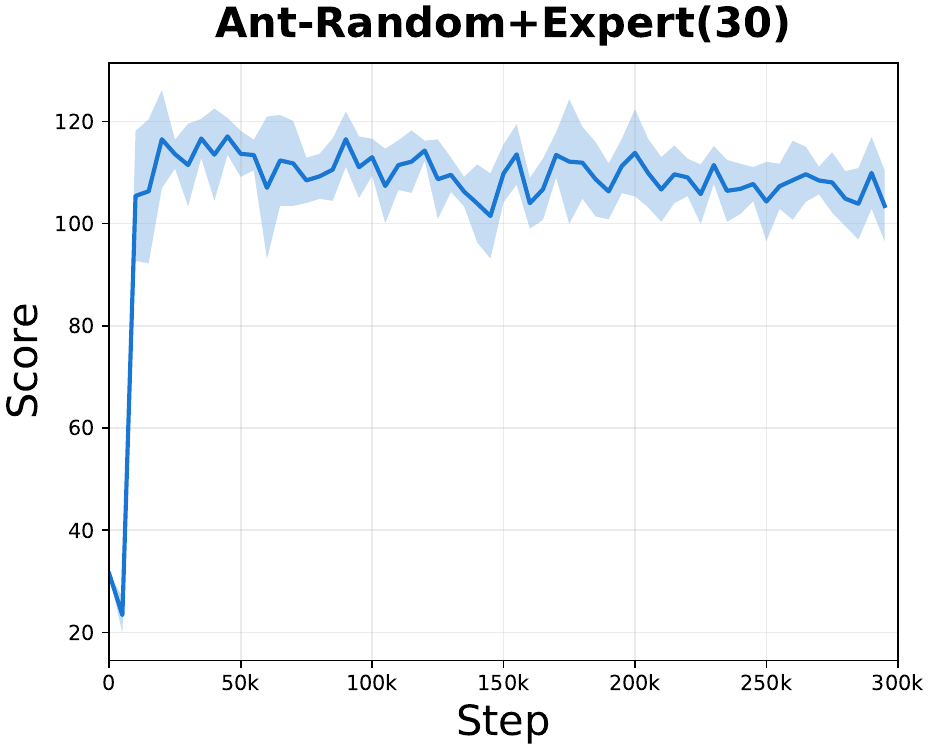}
        \caption{Ant-Random}
    \end{subfigure}
    
    \begin{subfigure}{0.24\textwidth}
        \includegraphics[width=\linewidth]{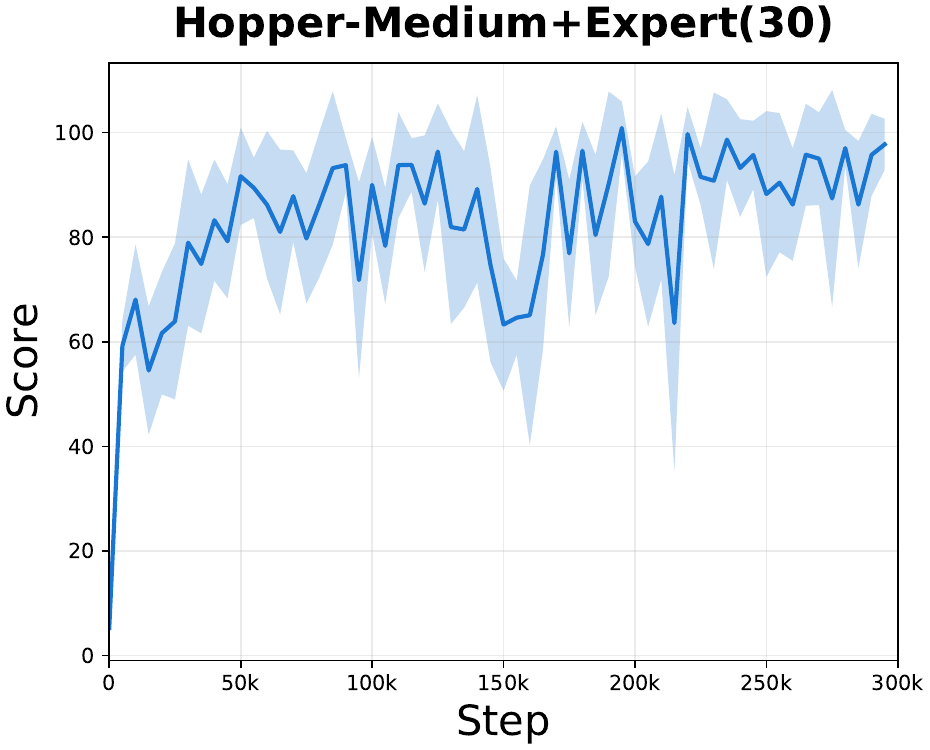}
        \caption{Hopper-Medium}
    \end{subfigure}
    \hfill
    \begin{subfigure}{0.24\textwidth}
        \includegraphics[width=\linewidth]{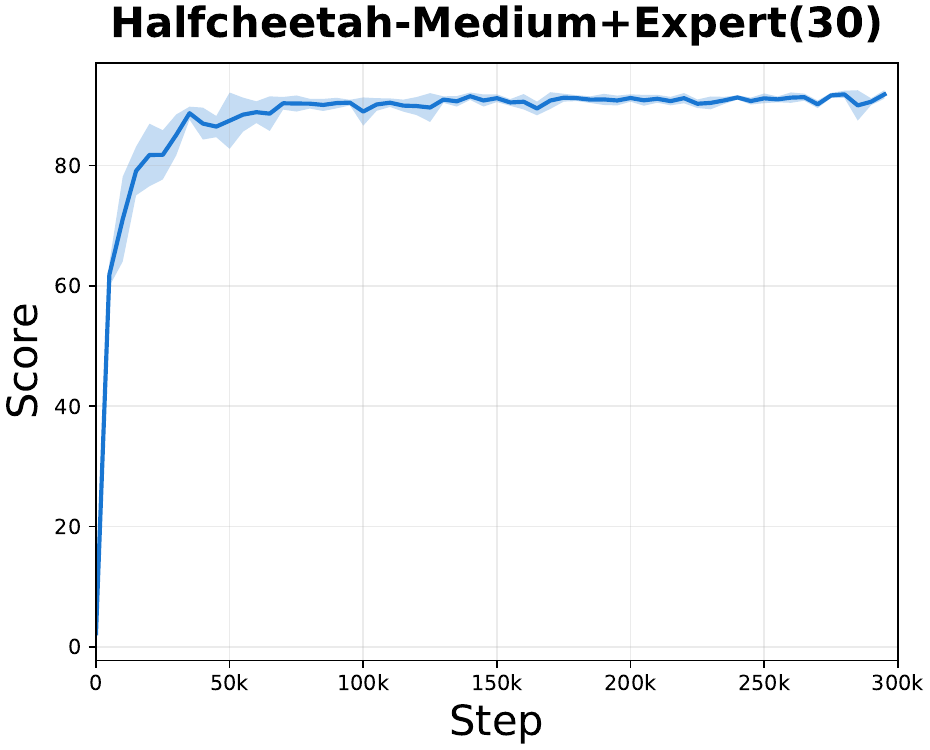}
        \caption{HalfCheetah-Medium}
    \end{subfigure}
    \hfill
    \begin{subfigure}{0.24\textwidth}
        \includegraphics[width=\linewidth]{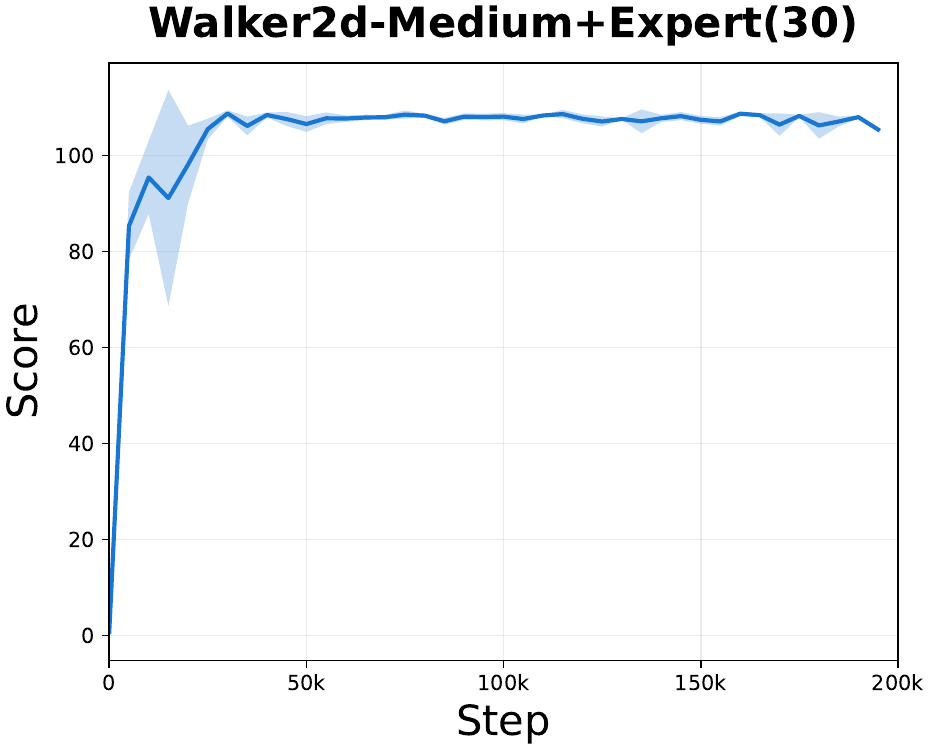}
        \caption{Walker2d-Medium}
    \end{subfigure}
    \hfill
    \begin{subfigure}{0.24\textwidth}
        \includegraphics[width=\linewidth]{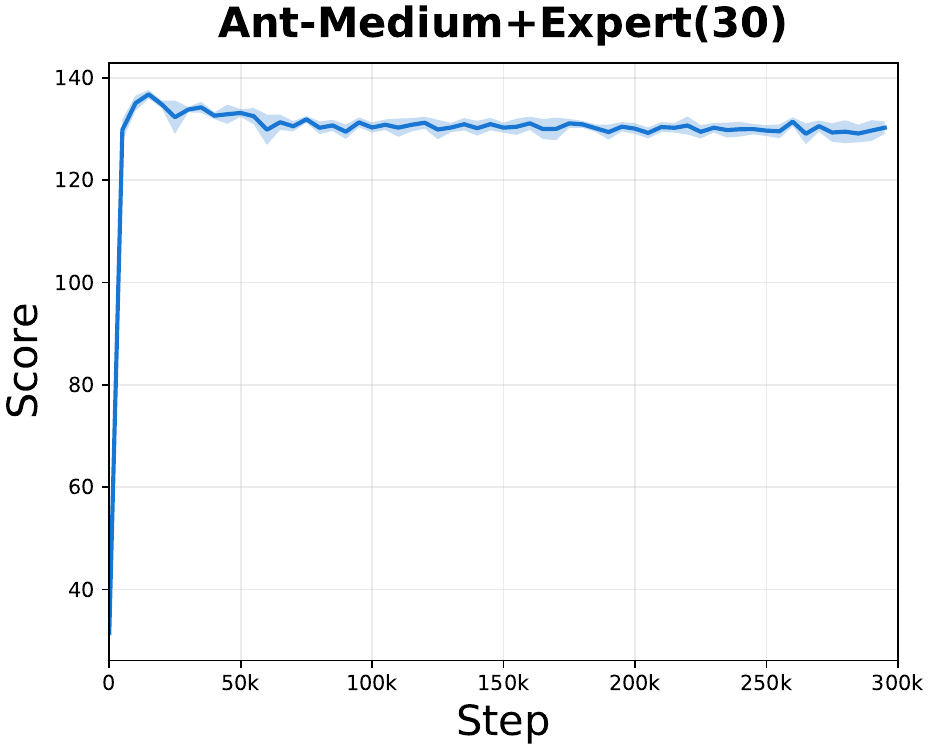}
        \caption{Ant-Medium}
    \end{subfigure}

    \caption{\textbf{Training Dynamics of TGE + ReBRAC.} The curves display the mean normalized score and standard deviation (shaded region). Similarly, the results demonstrate that the agent achieves rapid convergence to expert-level performance and maintains asymptotic stability, effectively robust to the noise in suboptimal datasets.}
    \label{fig:rebrac_curves}
\end{figure}

We provide the training curves of TGE combined with IQL and ReBRAC as downstream offline RL methods across different environments to demonstrate the training stability of our proposed method. Figure~\ref{fig:iql_curves} and Figure~\ref{fig:rebrac_curves} visualize the normalized D4RL score (y-axis) over training steps (x-axis) for the IQL and ReBRAC backbones, respectively.

\subsection{Comparison with baselines under support mismatch}
\begin{figure}[H]
    \centering
    \includegraphics[width=0.8\linewidth]{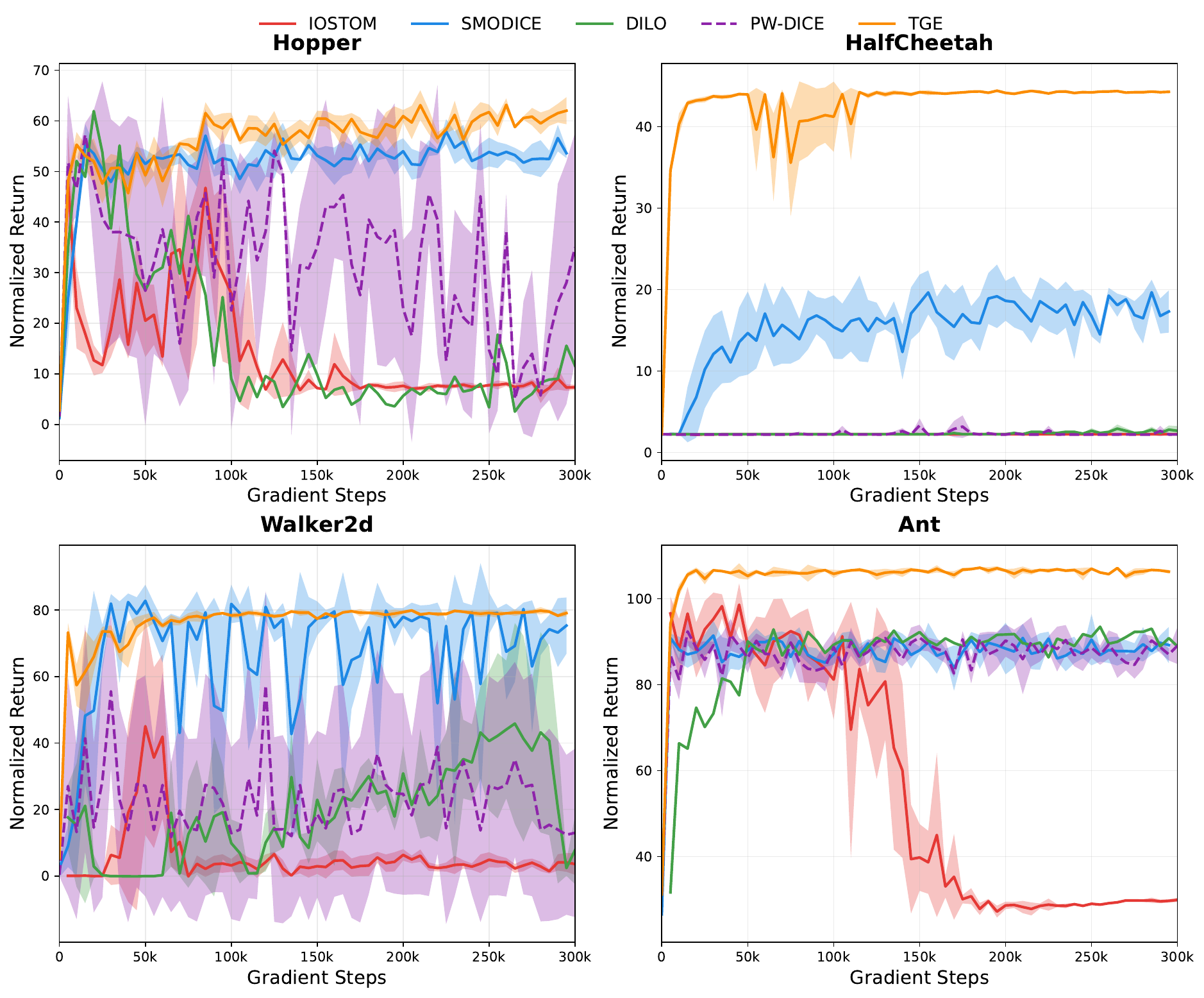}
    \caption{\textbf{Training dynamics under support mismatch.}
    Comparison of TGE with offline LfO baselines on the medium+random setting. TGE maintains strong performance across tasks.}
    \label{fig:training_dynamics_medrand}
\end{figure}

To complement the comparison in Table~\ref{tab:mismatched-result}, we visualize the training dynamics of TGE against offline LfO baselines on the medium+random setting, where the suboptimal dataset contains no expert trajectories. Figure~\ref{fig:training_dynamics_medrand} shows that TGE achieves better return and more stable learning compared to other methods across all four locomotion tasks. 

\newpage


\end{document}